\documentclass[dvipsnames,11pt]{article}

\usepackage{fancyhdr}
\usepackage{fullpage}
\usepackage{parskip}
\usepackage[utf8]{inputenc} 
\usepackage[T1]{fontenc}    
\usepackage[colorlinks = true,
            linkcolor = blue,
            urlcolor  = blue,
            citecolor = RoyalBlue,
            anchorcolor = blue]{hyperref}
\usepackage[utf8]{inputenc} 
\usepackage[T1]{fontenc}    
\usepackage{url}            
\usepackage{booktabs}       
\usepackage{amsfonts}       
\usepackage{nicefrac}       
\usepackage{microtype}      
\usepackage{placeins}
\usepackage{comment}
\usepackage[table]{xcolor}
\usepackage{xspace}
\usepackage{graphicx}
\usepackage{subcaption}
\usepackage{amsmath,amssymb}
\usepackage{multirow}

\usepackage{cleveref}   
\usepackage{tcolorbox}

\usepackage{algorithm}
\usepackage[noend]{algpseudocode}
\algblock{ParFor}{EndParFor}
\algnewcommand\algorithmicparfor{\textbf{parallel for}}
\algnewcommand\algorithmicpardo{\textbf{do}}
\algnewcommand\algorithmicendparfor{\textbf{end\ parallel for}}
\algrenewtext{ParFor}[1]{\algorithmicparfor\ #1\ \algorithmicpardo}
\algrenewtext{EndParFor}{\algorithmicendparfor}

\usepackage{natbib}
\PassOptionsToPackage{authoryear, sort&compress, round}{natbib} 

\definecolor{googleblue}{HTML}{8ab4f8}
\definecolor{googlered}{HTML}{f28b82}
\definecolor{googleyellow}{HTML}{fdd663}
\definecolor{googlegreen}{HTML}{81c995}

\newcommand{\dlc}{DiLoCo\xspace}
\newcommand{\fedopt}{FedOpt\xspace}

\newcommand{\dpfull}{Data-Parallel\xspace}

\newcommand{\tabemphbest}[1]{\cellcolor{blue!14}\textbf{#1}}
\newcommand{\tabemphgood}[1]{\cellcolor{blue!7}\textbf{#1}}

\newcommand{\tabemph}[1]{{#1}}  

\usepackage{pifont}
\newcommand{\cmark}{\color{ForestGreen}\ding{51}}
\newcommand{\xmark}{\color{Maroon}\ding{55}}

\newcommand{\eps}{\varepsilon}
\newcommand{\whigh}{W_{\text{high}}}
\newcommand{\epshigh}{\eps_{\text{high}}}

\newcommand{\wmed}{W_{\text{med}}}
\newcommand{\epsmed}{\eps_{\text{med}}}

\newcommand{\wlow}{W_{\text{low}}}
\newcommand{\epslow}{\eps_{\text{low}}}

\title{Communication-Efficient Language Model Training Scales Reliably and Robustly:\\ \textbf{Scaling Laws for DiLoCo}}

\author{
    Zachary Charles$^1$\footnote{Correspondence to \texttt{\url{zachcharles@google.com}}.}\hspace{5mm}
    Gabriel Teston$^2$\hspace{5mm}
    Lucio Dery$^3$\hspace{5mm}
    Keith Rush$^1$\hspace{5mm}
    \\
    Nova Fallen$^1$\hspace{5mm}
    Zachary Garrett$^1$\hspace{5mm}
    Arthur Szlam$^3$\hspace{5mm}
    Arthur Douillard$^3$
    \\
    \\\
    \textsuperscript{1}Google Research
    \hspace{4mm}
    \textsuperscript{2}Google Search
    \hspace{4mm}
    \textsuperscript{3}Google DeepMind
}
\date{}

\begin{document}

\maketitle

\begin{abstract}
As we scale to more massive machine learning models, the frequent synchronization demands inherent in data-parallel approaches create significant slowdowns, posing a critical challenge to further scaling.
Recent work~\citep{douillard2023diloco, douillard2025streaming} develops an approach (DiLoCo) that relaxes synchronization demands without compromising model quality. However, these works do not carefully analyze how DiLoCo's behavior changes with model size.
In this work, we study the scaling law behavior of DiLoCo when training LLMs under a fixed compute budget. We focus on how algorithmic factors, including number of model replicas, hyperparameters, and token budget affect training in ways that can be accurately predicted via scaling laws.
We find that DiLoCo scales both predictably and robustly with model size. When well-tuned, \textbf{DiLoCo scales better than data-parallel training with model size}, and can outperform data-parallel training even at small model sizes.
Our results showcase a more general set of benefits of DiLoCo than previously documented, including increased optimal batch sizes, improved downstream generalization with scale, and improved evaluation loss for a fixed token budget.
\end{abstract}

\begin{tcolorbox}[colback=googleblue, colframe=black, arc=4pt, boxsep=0.pt]%
\textbf{Harder:} DiLoCo's hyperparameters are robust and predictable across model scales.
\end{tcolorbox}%

\begin{tcolorbox}[colback=googleyellow, colframe=black, arc=4pt, boxsep=0.pt]%
\textbf{Better:} DiLoCo further improves over data-parallel training as model size increases.
\end{tcolorbox}%

\begin{tcolorbox}[colback=googlered, colframe=black, arc=4pt, boxsep=0.pt]%
\textbf{Faster:} DiLoCo uses orders of magnitude less bandwidth than data-parallel training.
\end{tcolorbox}%

\begin{tcolorbox}[colback=googlegreen, colframe=black, arc=4pt, boxsep=0.pt]%
\textbf{Stronger:} DiLoCo tolerates a significantly larger batch size than data-parallel training.
\end{tcolorbox}%

\section{Introduction}\label{sec:intro}
The default approach to training large language models (LLMs) continues to be large-batch distributed data-parallel training. However, bandwidth and communication constraints, which can be negligible at smaller scales, become dominant factors at larger scales. The frequent synchronization demands inherent in classical distributed approaches create significant slowdowns, posing a critical challenge to further scaling.  As a remedy, \citet{douillard2023diloco} propose DiLoCo (\textit{\underline{Di}stributed \underline{Lo}w-\underline{Co}mmunication}), a generalization of algorithms like Local SGD~\citep{mangasarian1993backpropagation,stich2018local} and FedAvg~\citep{mcmahan2017communication}, which enables training of LLMs in parallel across ``islands'' of compute (such as datacenters connected via low-bandwidth networks) by performing parallel training of models with only periodic synchronization. 

Unlike communication-reduction methods like quantization and sparsification, DiLoCo fundamentally alters training dynamics~\citep{rush2024drjax}. While \citet{douillard2023diloco, jaghouar2024opendiloco} show that \dlc yields comparable or better downstream evaluation metrics to data-parallel training at moderate model scales (up to 1.1 billion parameters), it is unclear how data-parallel training and DiLoCo compare at larger model scales. Moreover, \dlc has extra hyperparameters not present in data-parallel training that may be computationally prohibitive to tune at large enough scales.

This points to the need for DiLoCo {\it scaling laws}. Scaling laws generally give data-backed mechanisms for predicting facets of a model trained via a chosen algorithm (such as the evaluation loss after training on a given number of tokens)~\citep{kaplan2020scaling, hoffmann2022training}. We focus on two specific scaling laws: (1) predictions for evaluation loss as a function of model size and (2) predictions for optimal hyperparameter choices for a given model size (which can obviate the need to perform expensive hyperparameter tuning). In both cases, we are explicitly interested in how these compare to analogous scaling laws for data-parallel training.

\paragraph{Setting.} Throughout, we focus on the task of pre-training ``from scratch'' a model of size $N$ on some total number of tokens $D$. We are primarily concerned with predicting, for both data-parallel training and \dlc training, and as a function of $N$: (1) the evaluation loss $L$ after training, computed on a held-out set, and (2) optimal hyperparameter settings.

One path towards scaling laws for \dlc would be to develop them as a modification or functional transformation of scaling laws for data-parallel training. However, the facets of DiLoCo that are key to its communication-efficiency also make such a direct adaptation infeasible. First, DiLoCo operates by training $M$ models in parallel, with periodic synchronization every $H$ steps. The values of $M$ and $H$ depend on the ecosystem of compute available (such as the number of datacenters we are training a model over and the network bandwidth between them), and are absent in scaling laws for data-parallel training. Second, DiLoCo uses a bi-level optimization framework; each model replica performs data-parallel training, but upon synchronization we apply an ``outer'' optimization step~\citep{douillard2023diloco}. This means that \dlc has ``outer'' hyperparameters not present in \dpfull training that cannot be inferred from data-parallel hyperparameter scaling laws.

Instead, we develop scaling laws for data-parallel training and \dlc from the ground up, using similar methodology to work of \citet{kaplan2020scaling, hoffmann2022training}. Fixing the number of tokens $D$ to be the ``Chinchilla-optimal'' number of tokens~\citep{hoffmann2022training}, we model the evaluation loss and optimal hyperparameters of data-parallel training as functions of model size $N$, and we model the loss and optimal hyperparameters of DiLoCo as functions of model size $N$ and number of replicas $M$.\footnote{While we fix $H=30$ for these scaling laws, we also provide extensive ablations on the role of $H$ in \Cref{sec:sync_cadence}. This value of $H$ is large enough to imply a very substantial communication reduction, but (as our experiments show), \dlc with this setting of $H$ is still highly competitive with data-parallel training, sometimes even outperforming it.} We empirically estimate these functions using the final evaluation loss attained by models trained with both algorithms for varying hyperparameters (including learning rate, batch size, and ``outer'' learning rate for DiLoCo), model sizes (varying $N$ over 9 model sizes ranging from 35 million to 2.4 billion parameters), and numbers of DiLoCo replicas $M$.

\paragraph{Contributions.} Our core contribution are scaling laws for evaluation loss and optimal hyperparameters for both data-parallel training and \dlc based on the data described above. We show that these scaling laws provide good estimates of loss and optimal hyperparameters when extrapolating to larger model sizes.
To our surprise, these scaling laws predict that in many settings, the more communication-efficient \dlc algorithm would actually yield better evaluation loss than data-parallel training for the same token budget. Utilizing our scaling laws to predict the hyperparameters for \dlc, we tested these predictions when training models with 4 billion and 10 billion parameters. The scaling laws proved accurate, with \dlc outperforming data-parallel training as predicted, even while reducing total communication by a factor of over 100.

Our large amount of experimental data also enables us to analyze these algorithms at a deeper level, including an examination of their system characteristics. For each experiment, we provide an idealized end-to-end wall-clock training time under networks of varying bandwidth and latency.  We show that \dlc incurs a variety of benefits in comparison to data-parallel training, including (1) increased optimal batch size, allowing for greater horizontal scalability, (2) greater reductions in evaluation loss as model size increases, and (3) significantly less wall-clock training time.

One potentially surprising artifact of this data: \dlc improves training even when communication is not a bottleneck. Our experiments include \dlc with $M = 1$.
This algorithm, effectively an enhanced version of the Lookahead optimizer~\citep{zhang2019lookahead}, does not incur any communication reduction. However, it actually does better than data-parallel training across model sizes in terms of both evaluation loss and tolerance for larger batch sizes, via its use of an infrequent momentum operation.
Notably, \dlc, $M=1$ outperforms data-parallel training on both evaluation loss and training time. We show that \dlc with $M =1$ achieves lower evaluation loss at all model scales, and is more robust to larger batch sizes, greatly reducing wall-clock training time.

\section{Preliminaries}\label{sec:prelim}

Throughout, we let $\theta$ denote the model parameters. We let $\theta^{(t)}$ denote the model at step $t$. Since DiLoCo operates on $M$ parallel model parameters, we will use subscript notation $\theta_m$ to denote the $m$-th model. When there is no subscript, the parameters are assumed to be replicated across all DiLoCo replicas. For a batch of data $x$, we let $f(\theta, x)$ denote the loss of $\theta$ on the batch of data.

\begin{table}[htb]
    \centering
    \begin{minipage}{0.4\linewidth}
        \centering
        \caption{General Notation}
        \label{table:alg_independent_notation}
        \begin{tabular}{c|c}
            \toprule
            Symbol & Meaning \\
            \midrule
            $\theta$ & Model weights \\
            $N$ & Model size \\
            $L$ & Evaluation loss \\
            $T$ & Training steps \\
            $D$ & Token budget \\
            $C$ & Total FLOPs \\
            \bottomrule
        \end{tabular}
    \end{minipage}
    \begin{minipage}{0.58\linewidth}
        \caption{Algorithm-Specific Notation}
        \label{table:alg_specific_notation}
        \centering
        \begin{tabular}{c|cc}
        \toprule
        Symbol     & Data-Parallel     & DiLoCo \\
        \midrule
        $\gamma$ & Learning rate & Inner learning rate \\
        $\eta$ & -- & Outer learning rate \\
        $B$ & Batch size & Global batch size \\
        $M$ & -- & DiLoCo replicas \\
        $H$ & -- & Synchronization cadence \\
        \bottomrule
        \end{tabular}
    \end{minipage}
\end{table}

\begin{figure*}[ht!]
    \centering
    \includegraphics[width=0.9\linewidth]{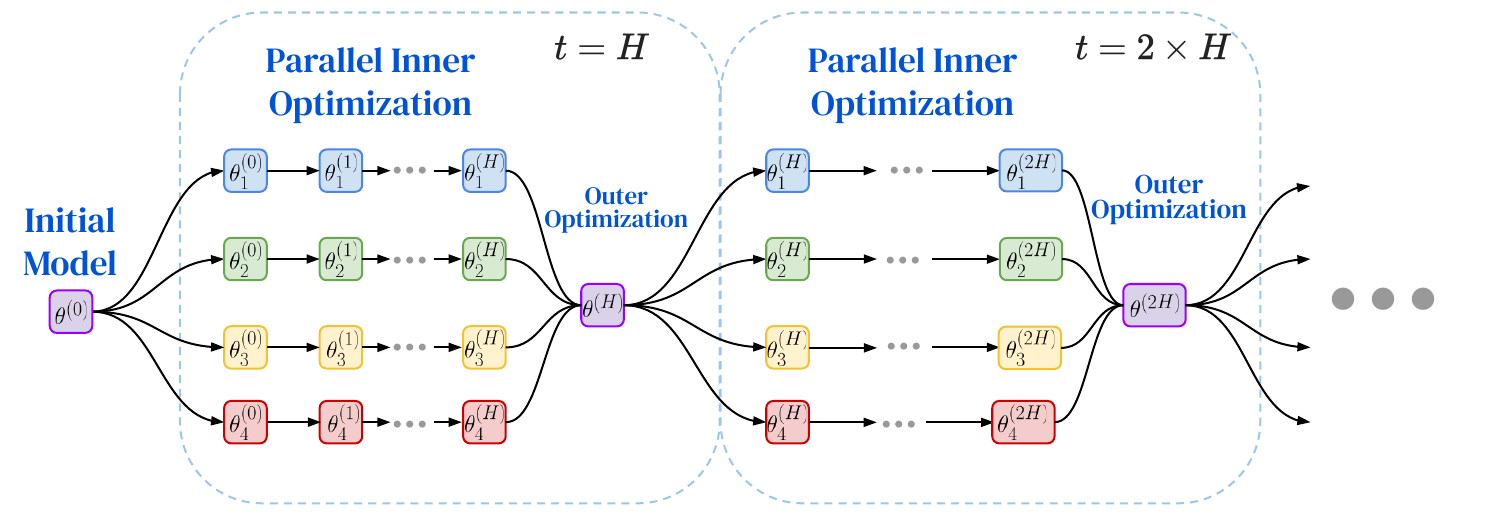}
    \caption{\textbf{DiLoCo}. Each DiLoCo model replica trains independently for $H$ inner optimization steps. These models are synchronized via an outer optimization step, usually involving momentum across outer optimization steps. In this figure, there are $M=4$ replicas.}
\label{fig:diloco}
\end{figure*} 

\subsection{DiLoCo: Distributed Low-Communication}\label{sec:diloco}

DiLoCo~\citep{douillard2023diloco} is a technique designed for training models in the presence of communication constraints. It is especially motivated by the training of large models across devices that are not all connected by low-latency bandwidth. To avoid incurring latency costs, DiLoCo trains $M$ models in parallel (ideally, training each one with co-located compute connected via low latency bandwidth), only synchronizing the models every $H$ steps. This is similar to the \fedopt algorithm used in federated learning~\citep{reddi2021adaptive}, but with the important difference that the replicas maintain their inner optimizer state across rounds.

In more detail, DiLoCo applies a bi-level optimization framework across multiple models: each DiLoCo replica has its own model $\theta_m^{(t)}$, and there is a global model $\theta^{(t)}$. At every step, each replica takes an \emph{inner} optimization step (\texttt{InnerOpt}). Every $H$ steps, each replica computes the $\Delta_m^{(t)} = \theta^{(t-H)} - \theta_m^{(t)}$, the difference in parameter space between the replica's current model and the most recently updated global model. We average these deltas across replicas, resulting in a vector $\Delta^{(t)}$ which we refer to as an \emph{outer gradient}. We treat this as a gradient estimate of the outer model\footnote{While empirically useful, it is worth noting that this operation is not completely grounded theoretically, as $\Delta^{(t)}$ is generally not a gradient of any function~\citep{pmlr-v167-charles22a}.} and an outer optimization step (\texttt{OuterOpt}) to the outer model $\theta^{(t - H)}$. This yields an updated outer model $\theta^{(t)}$ which is broadcast to all replicas and set as their current inner model. We give full pseudo-code for DiLoCo in \Cref{alg:diloco}.

\begin{algorithm}
\caption{DiLoCo} \label{alg:diloco}
\begin{algorithmic}[1]
\Require Loss function $f(\theta, x)$, batch size $B$, number of replicas $M$, synchronization cadence $H$.
\Require Initial model weights $\theta^{(0)}$, data shards $\{\mathcal{D}_1, \dots, \mathcal{D}_M\}$
\Require Optimizers $\texttt{InnerOpt}$ and $\texttt{OuterOpt}$
\State $\forall m$, $\theta_m^{(0)} \gets \theta^{(0)}$
\For{\texttt{step $t = 1 \ldots T$}}
    \ParFor{\texttt{replica $m = 1 \ldots M$}} 
    \State Receive a batch $x_m^{(t)} \sim \mathcal{D}_m$ of size $B / M$
    \State $g_m^{(t)} \gets \nabla_\theta f(\theta_m^{(t - 1)}, x_m^{(t)})$
    \State $\theta_m^{(t)} \gets \texttt{InnerOpt}(\theta_m^{(t-1)}, g_m^{(t)})$
\EndParFor
\item[]
\If{$t\bmod H = 0$}
    \State $\Delta^{(t)}_{m} \gets \theta^{(t-H)} - \theta_{m}^{(t)}$
    \State $\Delta^{(t)} \gets \frac{1}{M} \sum_{m=1}^M \Delta^{(t)}_{m}$
    \State $\theta^{(t)} \gets \texttt{OuterOpt}(\theta_m^{(t-H)}, \Delta^{(t)})$
    \State $\forall m, \theta^{(t)}_m \gets \theta^{(t)}$
\EndIf
\EndFor
\end{algorithmic}
\end{algorithm}

\subsection{\dpfull versus \dlc}

Throughout our work, we will perform model training via distributed data-parallel training, which we refer to as \dpfull for brevity, and \dlc. We discuss how they differ, as well as operational details we will use throughout. In \dpfull, at each step we take some batch of data of size $B$. In our work, like that of \citet{kaplan2020scaling, hoffmann2022training}, the batch size will refer to the number of tokens in a batch (as opposed to the number of sequences). We then compute a batch gradient and apply an optimization with a learning rate of $\gamma$.

In \dlc, at each step $t$, we take a global batch of data of size $B$, and evenly partition it at the sequence level across the $M$ DiLoCo replicas. Thus, the \emph{global} batch size is $B$, but each of the DiLoCo replicas uses a local batch of size $B/M$. Similarly to \dpfull, each replica then computes a batch gradient and applies an inner optimization step with a learning rate of $\gamma$. Unlike \dpfull, \dlc does outer optimization (on outer-gradients computed in parameter space) every $H$ steps with learning rate $\eta$.

An important comparison is \dpfull versus DiLoCo with $M = 1$. While similar, they are not identical. \dlc with $M = 1$ still has an outer optimizer step using \texttt{OuterOpt}, and is thus a variant of the Lookahead optimizer~\citep{zhang2019lookahead}. In our notation, ~\citep{zhang2019lookahead} corresponds to the setting when \texttt{OuterOpt} is SGD with learning rate $\eta$. In DiLoCo \texttt{OuterOpt} is often set to SGD with Nesterov momentum~\citep{douillard2023diloco} in which case \dlc with $M = 1$ becomes a variant of \dpfull with a momentum operation only applied every $H$ steps.\footnote{This is similar in spirit to the fast- and slow-momentum steps in AdEMAMix~\citep{pagliardini2024ademamix}, but yields different training dynamics since it uses a gradient estimate computed by linearizing across multiple training steps.}

When comparing between \dpfull and \dlc, we ensure that the model size $N$ and total token budget $D$ is the same. To compute an evaluation loss $L$ on some held-out set, for \dpfull we use the current model, and for \dlc we use the most recent global model. We summarize the algorithm-independent notation in \Cref{table:alg_independent_notation} and the algorithm-specific notation in \Cref{table:alg_specific_notation}.

\section{Experimental Methodology}\label{sec:exp_setup}

\begin{table}[t]
    \centering
    \caption{\textbf{Model details.} We present the size, number of layers, layer dimensions, and token budgets for each model trained. For the larger models (4B and 10B) we use scaling laws to predict optimal hyperparameters, rather than performing extensive hyperparameter tuning.}
    \label{tab:model_sizes}
    \begin{tabular}{c|cccccc}
        \toprule
        Model & Transformer & Attention & QKV & Hidden & Token & Hyperparameter \\
        Scale & Layers & Heads & Dimension & Dimension & Budget & Sweep \\
        \midrule
        35M & 6 & 8 & 512 & 2,048 & 70M & \cmark \\
        90M & 9 & 12 & 768 & 3,072 & 1.8B & \cmark \\
        180M & 12 & 16 & 1,024 & 4,096 & 3.6B & \cmark \\
        330M & 15 & 20 & 1,280 & 5,120 & 6.6B & \cmark \\
        550M & 18 & 24 & 1,536 & 6,144 & 11B & \cmark \\
        1.3B & 24 & 32 & 2,048 & 8,192 & 26B & \cmark \\
        2.4B & 30 & 40 & 2,560 & 10,240 & 48B & \cmark \\
        4B & 36 & 48 & 3,072 & 12,288 & 80B & \xmark \\
        10B & 48 & 64 & 4,096 & 16,384 & 200B & \xmark \\
        \bottomrule
    \end{tabular}
\end{table} 

\paragraph{Model architecture.} We use a Chinchilla-style decoder-only transformer architecture~\citep{hoffmann2022training}. As suggested by \citet{wortsman2023small} and \citet{jaghouar2024opendiloco}, we use QK-LayerNorm to reduce sensitivity to learning rate. We also use z-loss regularization~\citep{chowdhery2023palm} to increase training stability. We use a vocabulary size of 32,768: 32,000 in-vocabulary words, and extra tokens for BOS and out-of-vocabulary (extended so that the vocabulary size is a power of 2, which is useful for sharding computations across accelerators). We pack multiple sequences into each batch, with a maximum sequence length of 2,048 throughout. We train all of our models from scratch, as we are primarily interested in scaling laws for pre-training regimes in this work.

We train on a family of models varying the number of transformer layers, number of attention heads, QKV dimension, and feed-forward layer hidden dimension. Details on the architecture for each model scale are given in \Cref{tab:model_sizes}. As we will discuss in \Cref{sec:scaling_law_exps}, we use the Chinchilla token budget unless otherwise noted and do extensive hyperparameter sweeps on all models except the two largest (4B and 10B).

\paragraph{Datasets.} In most experiments, we train our models using the train split of the C4 dataset~\citep{raffel2020exploring} throughout. We report evaluation metrics on C4's held-out validation set. Additionally, we compute downstream zero-shot evaluation metrics on 3 tasks: HellaSwag~\citep{zellers2019hellaswag}, Piqa~\citep{bisk2020piqa}, and Arc-Easy~\citep{clark2018think}. When performing overtraining ablations, we use the Dolma dataset~\citep{soldaini2024dolma}.

\paragraph{Algorithms and optimizers.} We use AdamW~\citep{loshchilov2017decoupled} as the optimizer for \dpfull and the inner optimizer for \dlc. For both algorithms, we set $\beta_1 = 0.9$ and $\beta_2 = 0.99$. We do 1000 steps of warmup followed by cosine learning rate decay. Following \citep{wang2024set}, we set the weight decay parameter $\lambda$ to $T^{-1}$ where $T$ is the total number of training \emph{steps} (which crucially depends on the batch size and token budget). We decay to 5\% of the peak learning rate by the end of training. For training stability, we clip (inner) gradients to a global $\ell_2$ norm of $1$. We do not clip outer gradients. For \dlc, we use SGD with Nesterov momentum \citep{sutskever2013nesterov} as the outer optimizer, as suggested by~\citet{douillard2023diloco}. We use a momentum term of 0.9, and a constant outer learning rate. Unless otherwise specified, we set $H = 30$.

\paragraph{Implementation.} We use a modified version of NanoDO~\citep{nanodo} that uses DrJAX~\citep{rush2024drjax} to parallelize inner training steps across replicas and surface the model replica axis for explicit programming. This was crucial for better scaling performance in JAX~\citep{jax2018github}, as DrJAX provides an enriched version of \texttt{jax.vmap} that provides more explicit sharding information about the DiLoCo replicas. The outer optimization is implemented using an all-reduce. Data-parallel training is implemented as a special case with a single model replica, and no outer optimization step. We use bfloat16 representation of model weights and gradients throughout. We perform most of our experiments on TPUv5e and TPUv6e, and for the largest scales on TPUv5.

\paragraph{Idealized wall-clock time.} For each experiment, we also compute an idealized wall-clock time for training, that factors in both an idealized computation time and idealized communication time. We specifically measure the end-to-end wall-clock time (sometimes referred to as elapsed real time). Greater horizontal parallelization, such as via doubling the batch size, will therefore reduce wall-clock time. Our model assumes that we are training a model across multiple datacenters. Within a datacenter, we have a high-bandwidth network. Across datacenters, we either have a high-, medium-, or low-bandwidth network. For the idealized communication time, we always use the high-bandwidth network for the within-datacenter network, and one of the three for the cross-datacenter network. For details on the idealized wall-clock time, see \Cref{sec:wallclock_model}.

\subsection{Scaling Law Experiments}\label{sec:scaling_law_exps}

We perform comprehensive hyperparameter sweeps for \dpfull and \dlc on models ranging from 35M to 2.4B We sweep over the learning rate $\gamma$ using integer powers of $\sqrt{2}$ and batch size $B$ using powers of $2$. For DiLoCo, we train using $M = 1, 2, 4, 8$ and sweep the outer learning rate $\eta$ over $\{0.2, 0.4, 0.6, 0.8, 1.0\}$. The initial grids for (inner) learning rate and batch size depend on the model size, and were extended as needed until the minimum loss value was obtained on an interior point in all hyperparameter grids.

Using this data, we derive scaling laws to predict evaluation loss and optimal hyperparameters for larger models. We use the predicted hyperparameters to train models with 4B and 10B parameters, in order to validate the scaling laws empirically. Following the Chinchilla scaling laws~\citep{hoffmann2022training} we assume that the optimal token budget is given by $D = 20N$. Unless otherwise indicated, in all experiments we train using this number of tokens. This means that for a fixed model size, if we double the batch size $B$, we halve the number of training steps.

\begin{figure}[ht]
    \centering
    \begin{subfigure}[b]{0.45\linewidth}
    \includegraphics[width=\linewidth]{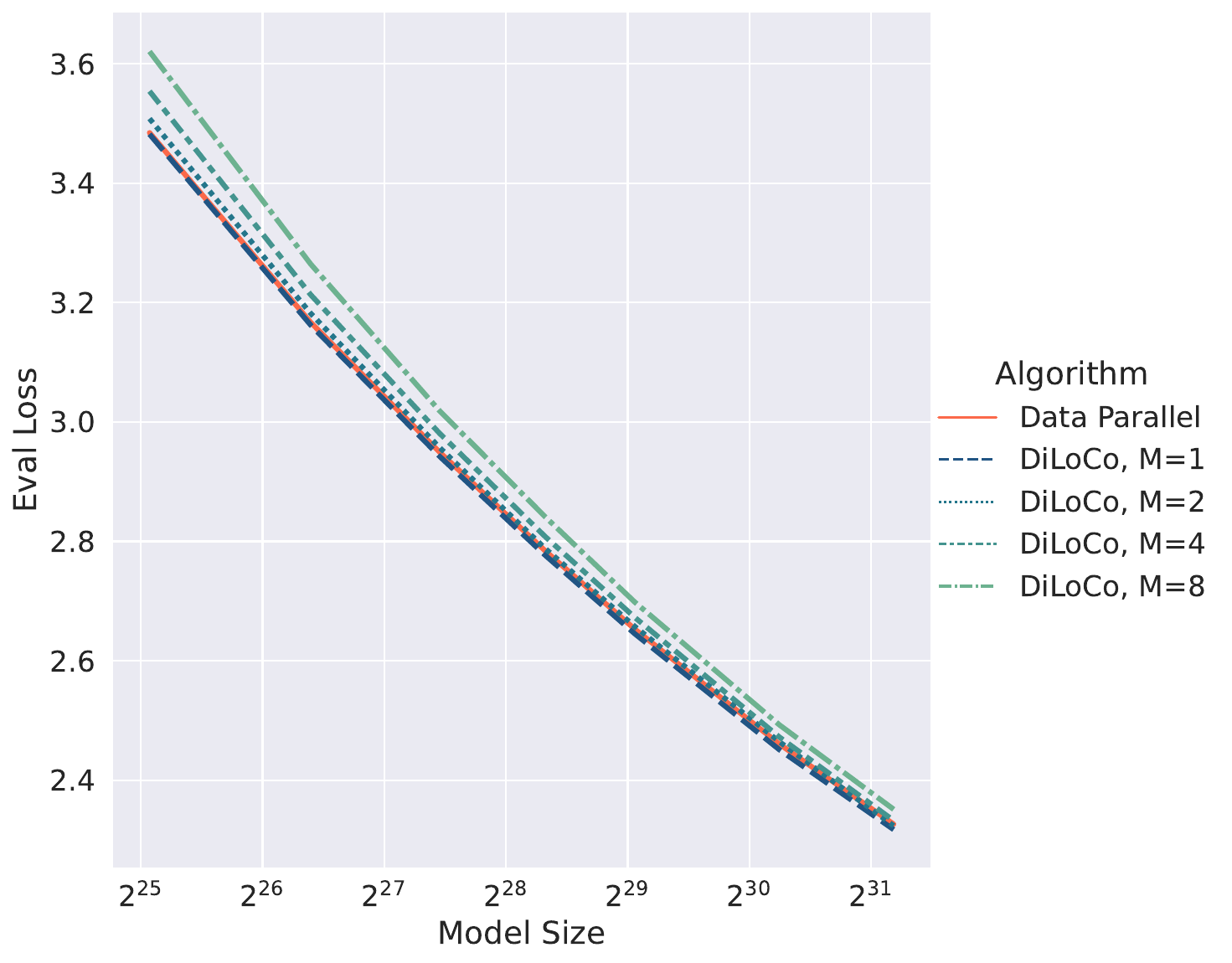}
    \caption{Evaluation loss for various algorithms, as a function of $N$.}
    \end{subfigure}
    \hspace{1cm}
    \begin{subfigure}[b]{0.45\linewidth}
    \includegraphics[width=\linewidth]{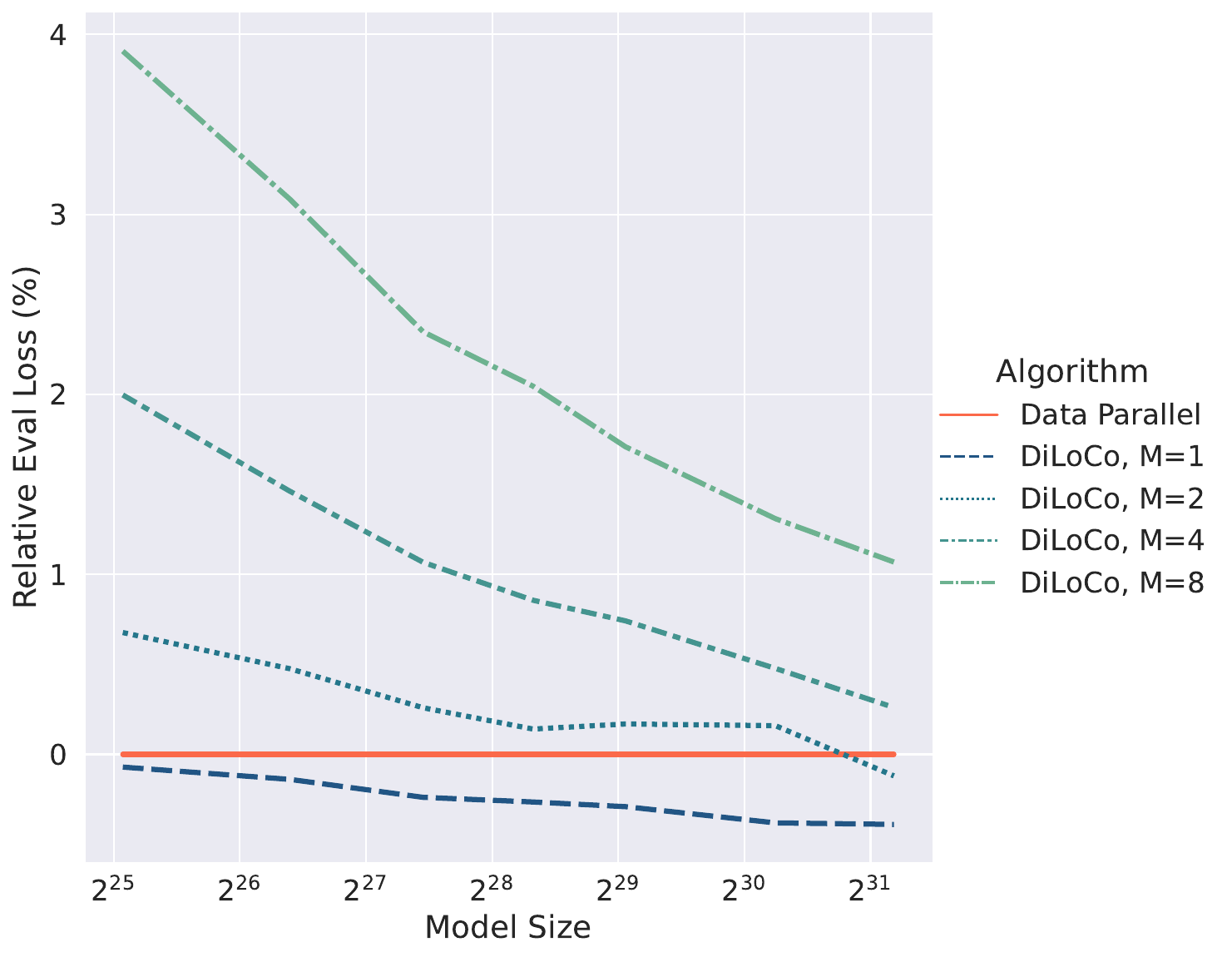}
    \caption{Percentage difference in evaluation loss, relative to \dpfull.}
    \end{subfigure}
    \caption{\textbf{DiLoCo does better with scale.} We compare Data-Parallel to DiLoCo for varying model sizes $N$. For all $M$, DiLoCo improves monotonically wrt \dpfull as $N$ increases.}
    \label{fig:best_eval_loss}
\end{figure}

\section{Empirical Findings}\label{sec:exp_results}

Before discussing our process for fitting the specific scaling laws, we talk about the empirical results and four critical findings that are worth highlighting independently.


\begin{tcolorbox}[colback=googlegreen, colframe=black, arc=4pt, boxsep=0.pt]%
\textbf{Finding 1: Scale.} DiLoCo's evaluation loss improves relative to \dpfull as $N$ increases (\Cref{fig:best_eval_loss} and \Cref{table:best_eval_loss}). Scaling laws predict \dlc with $M = 2$ achieves lower loss than \dpfull above several billion parameters, a phenomenon validated by both the largest model over which we performed tuning and our 4B and 10B model training runs (\Cref{table:scaling_extrapolation}).
\end{tcolorbox}%

\begin{table}[ht!]
    \centering
    \caption{\textbf{Evaluation loss for \dpfull and \dlc at varying model sizes}. We vary model size $N$ and number of replicas $M$, and report evaluation loss $L$ along with the percentage difference with respect to \dpfull (DP). We indicate settings where DiLoCo achieves a lower loss than \dpfull in bold.}
    \label{table:best_eval_loss}
    \begin{tabular}{cccccc}
        \toprule
        \multirow{2}{*}{$N$} & \multirow{2}{*}{DP} & \multicolumn{4}{c}{\dlc} \\
        \cmidrule{3-6}
        & & $M = 1$ & $M = 2$ & $M = 4$ & $M = 8$ \\
        \midrule
35M & $3.485$ & \tabemphgood{$\boldsymbol{3.482}$ $(\boldsymbol{-0.1\%})$} & $3.508$ $(+0.7\%)$ & $3.554$ $(+2.0\%)$ & $3.621$ $(+3.9\%)$ \\
90M & $3.167$ & \tabemphgood{$\boldsymbol{3.162}$ $(\boldsymbol{-0.1\%})$} & $3.182$ $(+0.5\%)$ & $3.213$ $(+1.5\%)$ & $3.265$ $(+3.1\%)$ \\
180M & $2.950$ & \tabemphgood{$\boldsymbol{2.943}$ $(\boldsymbol{-0.2\%)}$} & $2.957$ $(+0.3\%)$ & $2.981$ $(+1.1\%)$ & $3.019$ $(+2.3\%)$ \\
335M & $2.784$ & \tabemphgood{$\boldsymbol{2.777}$ $(\boldsymbol{-0.3\%})$} & $2.788$ $(+0.1\%)$ & $2.808$ $(+0.9\%)$ & $2.841$ $(+2.0\%)$ \\
550M & $2.653$ & \tabemphgood{$\boldsymbol{2.645}$ $(\boldsymbol{-0.3\%})$} & $2.657$ $(+0.2\%)$ & $2.673$ $(+0.7\%)$ & $2.698$ $(+1.7\%)$ \\
1.3B & $2.460$ & \tabemphgood{$\boldsymbol{2.451}$ $(\boldsymbol{-0.4\%})$} & $2.464$ $(+0.2\%)$ & $2.472$ $(+0.5\%)$ & $2.493$ $(+1.3\%)$ \\
2.4B & $2.326$ & \tabemphgood{$\boldsymbol{2.317}$ $(\boldsymbol{-0.4\%})$} & \tabemphgood{$\boldsymbol{2.323}$ $(\boldsymbol{-0.1\%})$} & $2.332$ $(+0.3\%)$ & $2.351$ $(+1.1\%)$ \\
        \bottomrule
    \end{tabular}
\end{table}

\begin{table}[ht]
    \centering
    \caption{\textbf{\dlc outperforms \dpfull at larger scales.} Here we show the evaluation results on 4B and 10B models, using hyperparameters predicted by scaling laws. We indicate settings where \dlc reaches lower loss than \dpfull in bold.}
    \label{table:scaling_extrapolation}
    \begin{tabular}{ccc}
        \toprule
        Algorithm & \multicolumn{2}{c}{Loss} \\
        \cmidrule{2-3}
        & 4B & 10B \\
        \midrule
        \dpfull & 2.224 & 2.090 \\
        \hline
        \dlc, $M = 1$ & \tabemphgood{2.219 (-0.22\%)} & \tabemphgood{2.086 (-0.19\%)} \\
        \hline
        \dlc, $M = 2$
                         & \tabemphgood{2.220 (-0.18\%)} & \tabemphgood{2.086 (-0.19\%)} \\
        \hline
        \dlc, $M = 4$ & 2.230 (+0.18\%) & 2.096 (+0.29\%)\\
        \bottomrule
    \end{tabular}
\end{table}

This finding has two separate, but related, components. First, as noted above, \dlc with $M = 1$ seems to attain lower evaluation loss than \dpfull on all model sizes. Moreover, the gap between \dpfull and \dlc, $M = 1$ widens as $N$ increases. Second, on most model sizes, \dlc with $M \geq 2$ achieves a higher evaluation loss. However, if we look at the signed percentage difference between \dlc and \dpfull, we see that as $N$ increases, \dlc does better and better relative to \dpfull, and outperforms \dpfull at $M=2$ when $N$ = 2.4B.

For example, we give the evaluation loss achieved by \dpfull and \dlc for each model size $N$ in \Cref{table:best_eval_loss}. We see that for all values of $M$, the percentage difference strictly decreases with $N$. DiLoCo with $M = 4$ goes from attaining 2\% higher evaluation loss for $N$ = 35M  parameters to only a 0.26\% higher evaluation loss at $N$ = 2.4B parameters. This same information is plotted in Figure \ref{fig:best_eval_loss}. We see that as $N$ increases, the relative evaluation loss of DiLoCo decreases.

We validated this by training 4B and 10B models with hyperparameters set via our scaling laws. Though~\Cref{fig:best_eval_loss} show to the `interpolation' regime, the result of extensive sweeps, these findings qualitatively carry over to the extrapolation regime, allowing us to train 4B and 10B models to lower evaluation losses using DiLoCo when $M=1,2$. \Cref{table:scaling_extrapolation} shows the results of training with our extrapolated hyperparameters, with deeper commentary on the extrapolation regime in \Cref{sec:scaling_laws}.

\begin{figure}[htb]
    \centering
    \begin{subfigure}[b]{0.48\linewidth}
    \includegraphics[width=\linewidth]{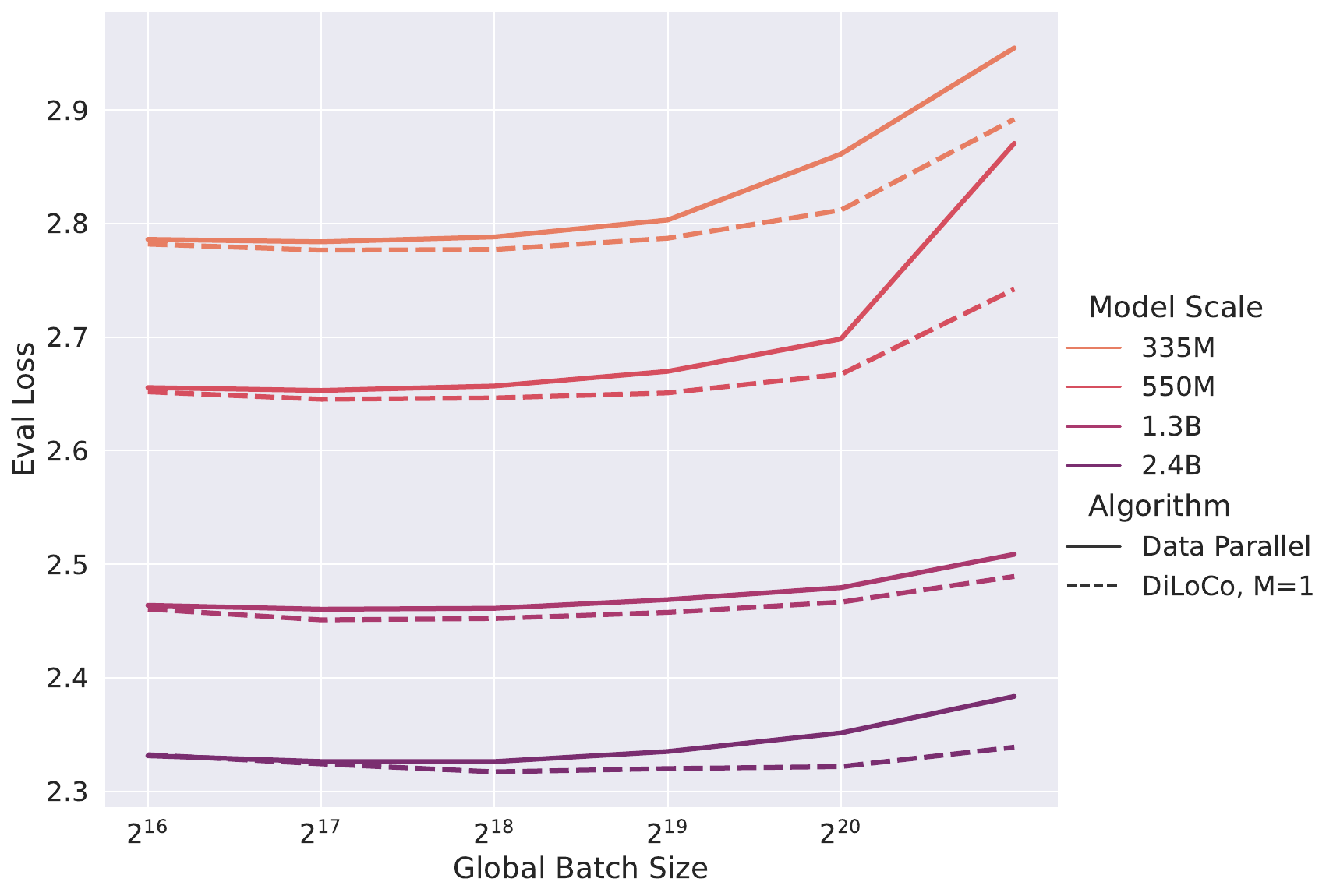}
    \caption{Evaluation loss.}
    \end{subfigure}
    \begin{subfigure}[b]{0.48\linewidth}
    \includegraphics[width=\linewidth]{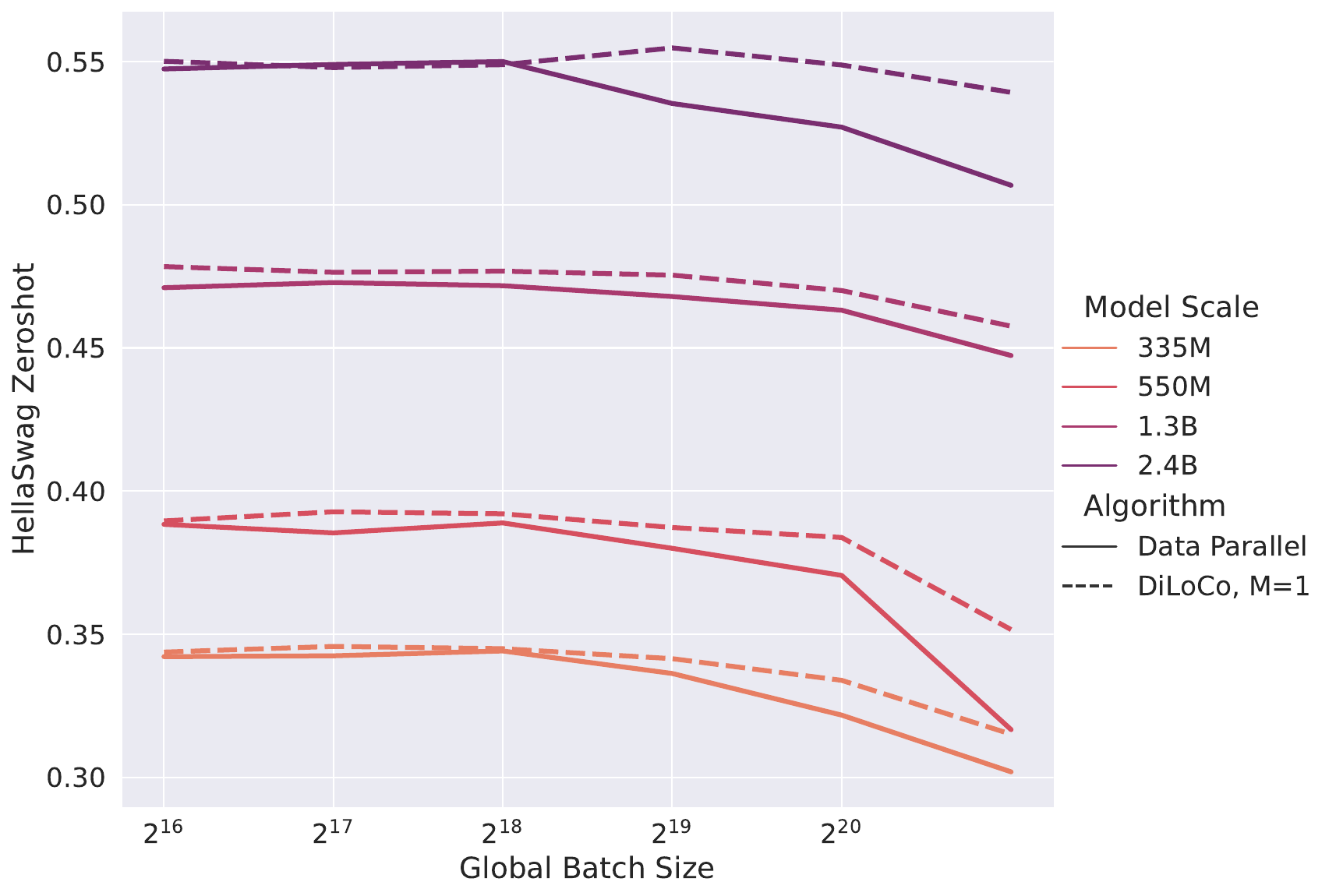}
    \caption{Zero-shot accuracy on HellaSwag.}
    \end{subfigure}
    \caption{\textbf{DiLoCo with $M = 1$ generalizes better than \dpfull}. We present the evaluation loss and downstream accuracy of \dpfull and DiLoCo with $M = 1$ for varying model and global batch sizes (measured in tokens). In all settings, DiLoCo with $M = 1$ does better than \dpfull, and the gap between them increases with batch size. We see similar results for other model sizes, but omit for the sake of brevity.}
    \label{fig:compare_dp_m1_main}
\end{figure}

\begin{tcolorbox}[colback=googlegreen, colframe=black, arc=4pt, boxsep=0.pt]%
\textbf{Finding 2: Single Replica \dlc.} \dlc with $M=1$ attains lower evaluation loss than \dpfull across model scales (see \Cref{fig:compare_dp_m1_main}).
\end{tcolorbox}%

Another key findings is that in virtually all settings, DiLoCo with $M = 1$ attained lower evaluation loss and higher downstream zero-shot evaluation accuracy than \dpfull.
Moreover, the performance of DiLoCo with $M = 1$ exhibited much greater stability with respect to batch size; doubling or quadrupling the batch size greatly reduced performance of \dpfull, but had little effect on DiLoCo, $M = 1$, as depicted in \Cref{fig:compare_dp_m1_main}.

\begin{figure}[ht]
    \centering
    \includegraphics[width=\linewidth]{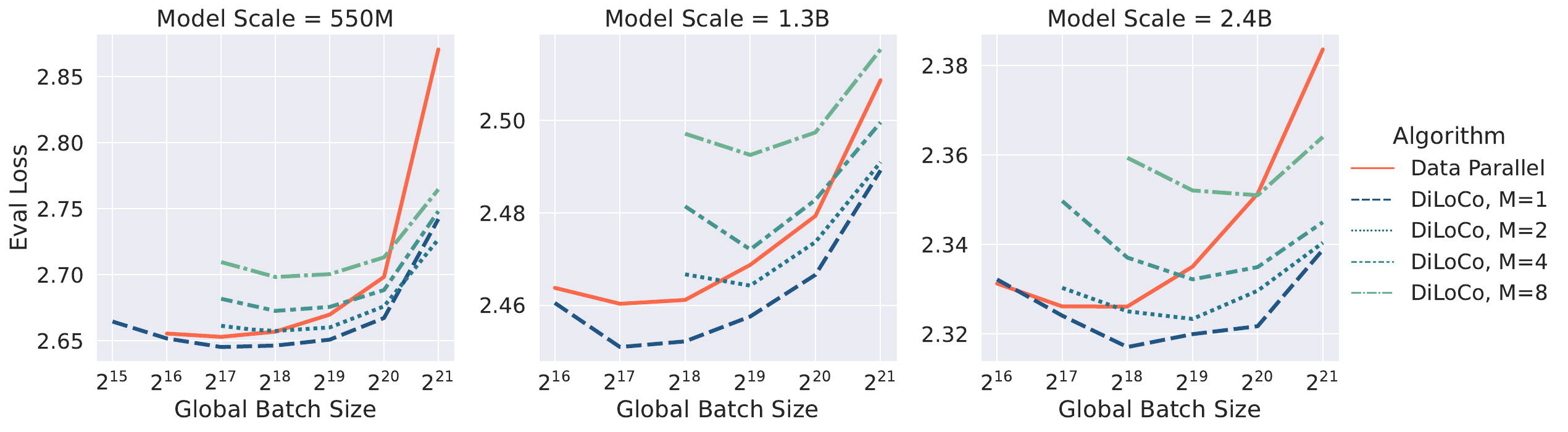}
    \caption{\textbf{DiLoCo increases optimal batch size, part 1.} Evaluation loss of Data-Parallel and DiLoCo as a function of global batch size (in tokens). For all $M$, DiLoCo exhibits larger optimal batch size than \dpfull. Moreover, the optimal batch size increases as a function of $M$. We see similar results for other model sizes, but omit for conciseness.}
    \label{fig:batch_size_eval_loss_reduced}
\end{figure}
\begin{figure}[htb]
    \centering
    \includegraphics[width=\linewidth]{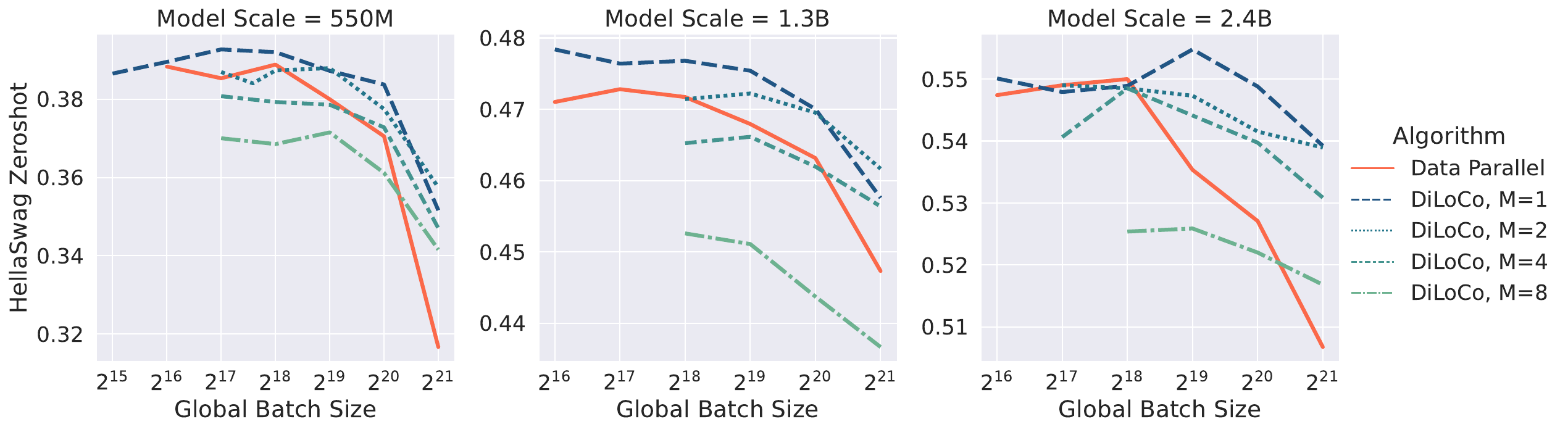}
    \caption{\textbf{DiLoCo increases optimal batch size, part 2.} Zero-shot accuracy on HellaSwag of Data-Parallel and DiLoCo as a function of global batch size (in tokens). Even at smaller model sizes, DiLoCo with $M = 2$ attains higher accuracy for larger global batch sizes. We see similar results for other model sizes, but omit for conciseness.}
    \label{fig:batch_size_hellaswag_reduced}
\end{figure}
\begin{figure}[thb]
    \centering
    \begin{subfigure}[b]{0.3\linewidth}
        \includegraphics[width=\linewidth]{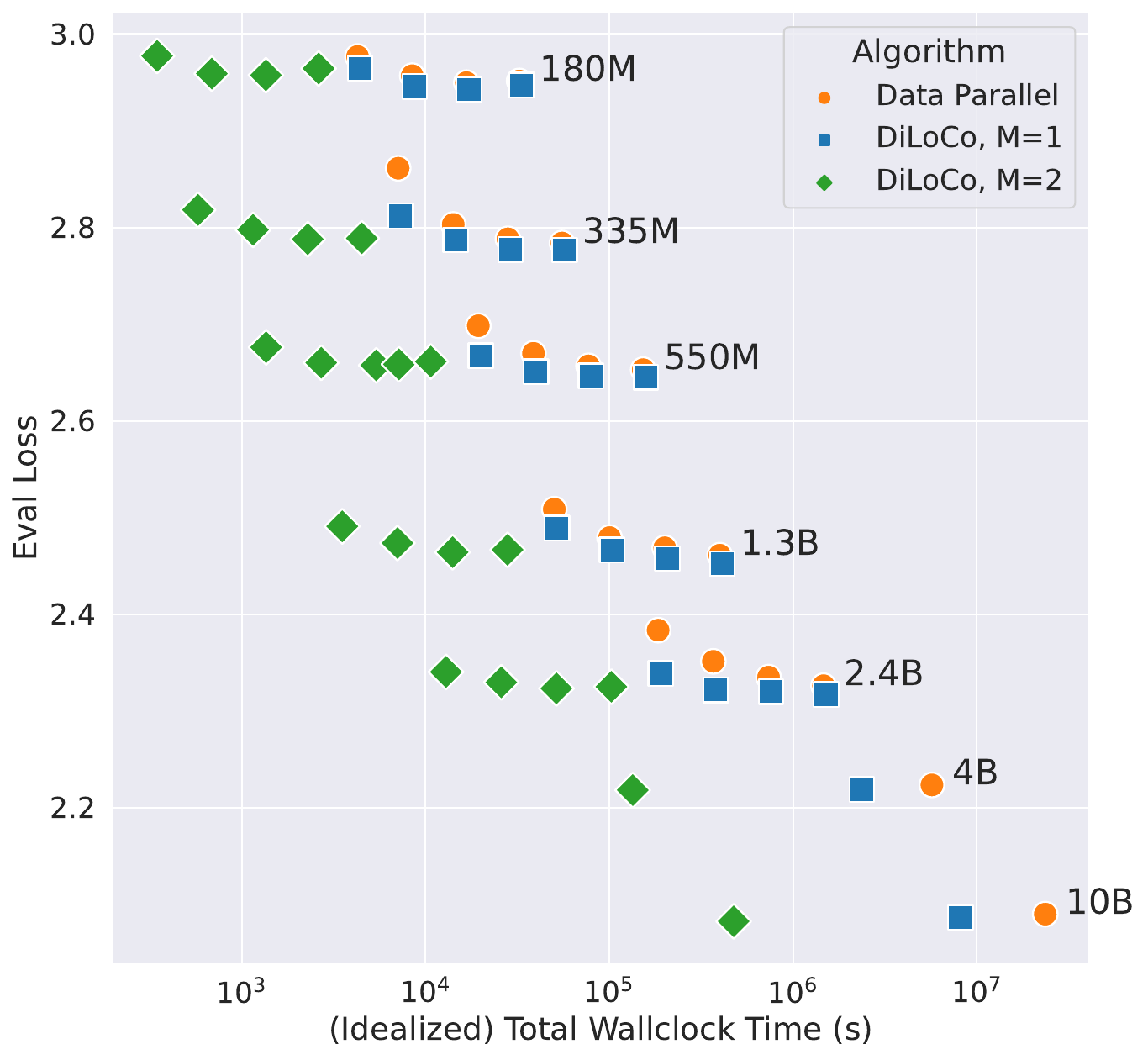}
        \caption{Network with a bandwidth of 10 gigabits/s and a latency of $10^{-2}$ seconds (\textbf{low-bandwidth}).}
    \end{subfigure}
    \hfill
    \begin{subfigure}[b]{0.3\linewidth}
        \includegraphics[width=\linewidth]{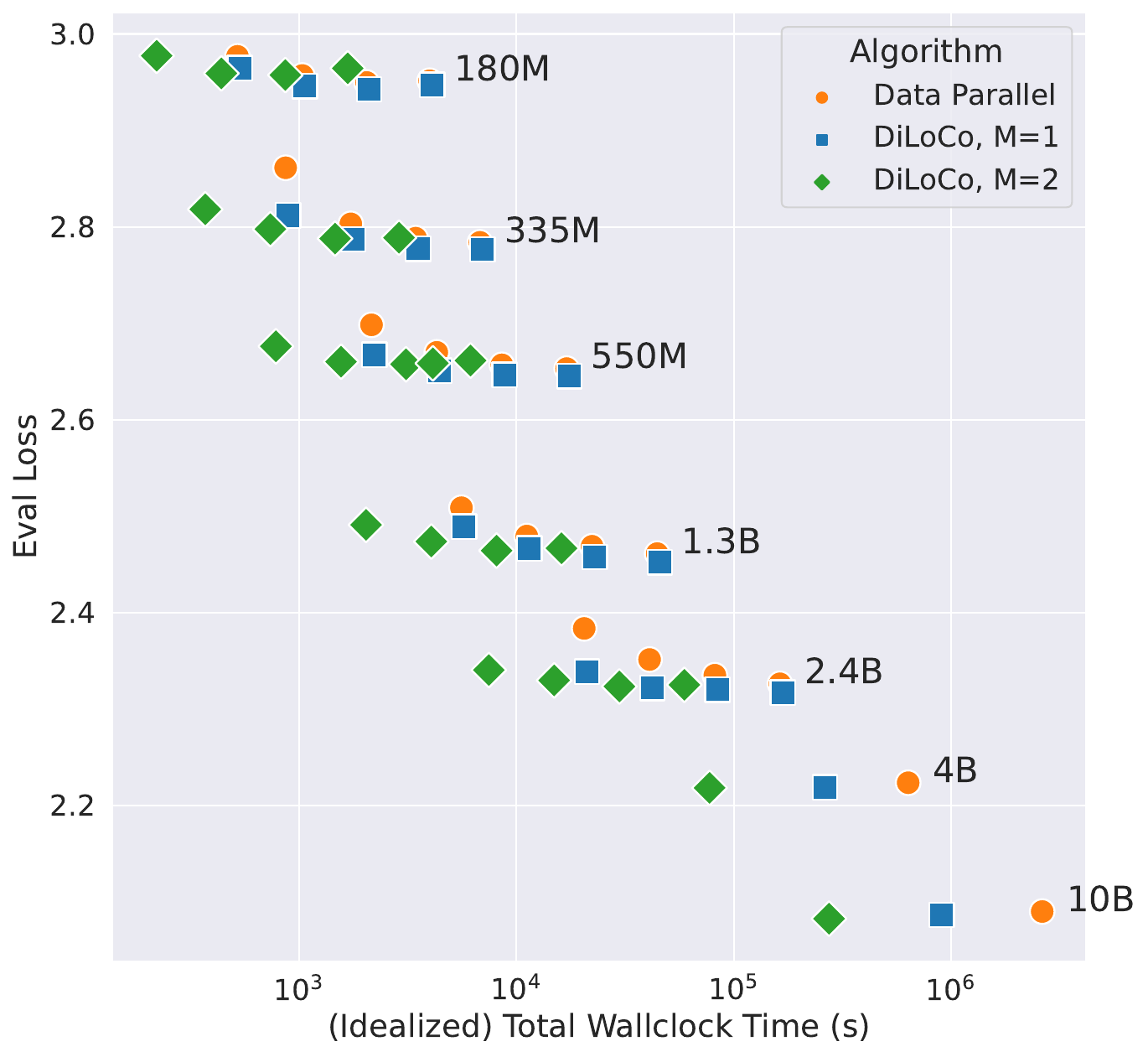}
        \caption{Network with a bandwidth of 100 gigabits/s and a latency of $10^{-3}$ seconds (\textbf{medium-bandwidth}).}
    \end{subfigure}
    \hfill
    \begin{subfigure}[b]{0.3\linewidth}
        \includegraphics[width=\linewidth]{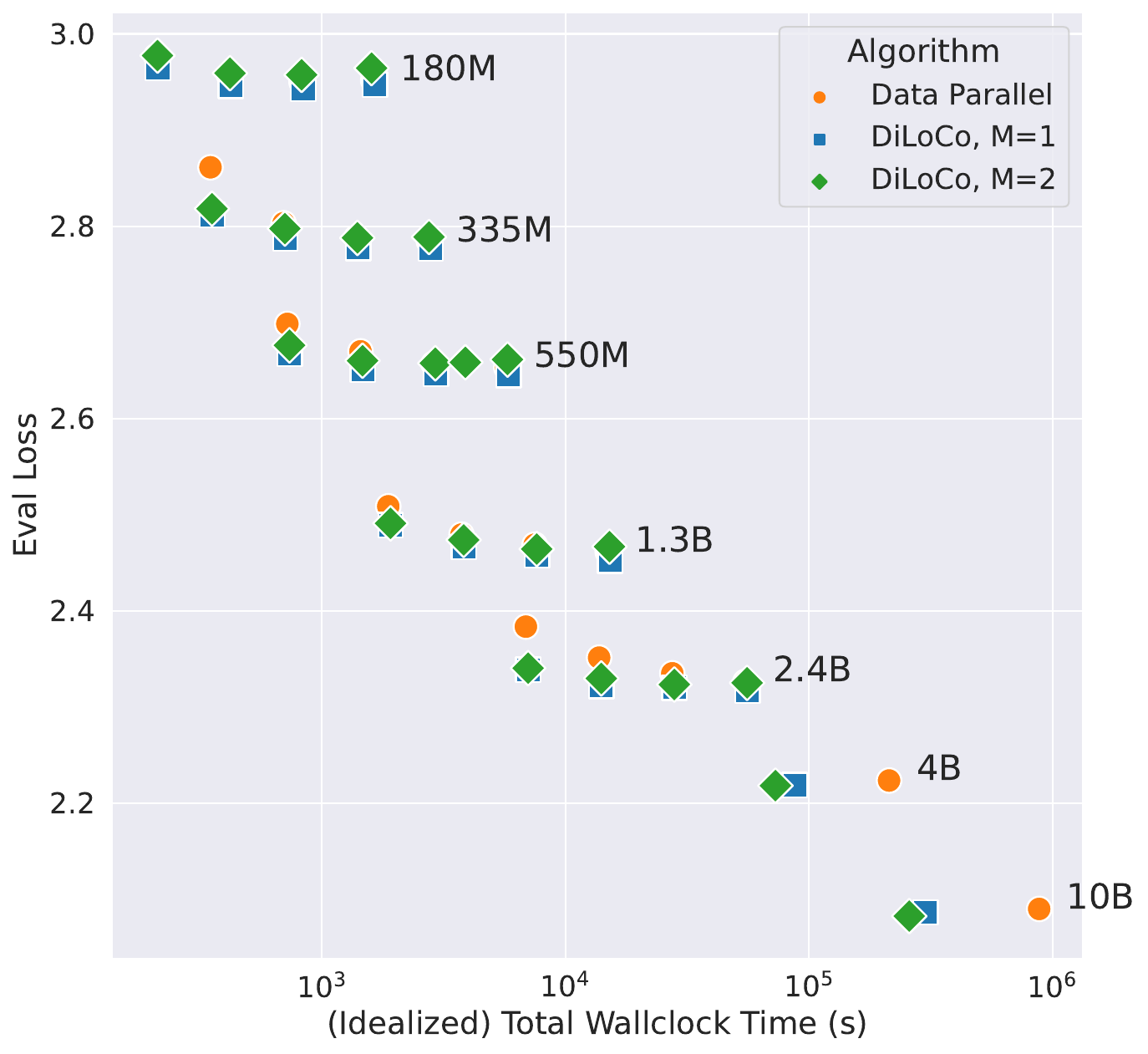}
        \caption{Network with a bandwidth of 400 gigabits/s and a latency of $10^{-4}$ seconds (\textbf{high-bandwidth}).}
    \end{subfigure}
    \caption{\textbf{DiLoCo trains faster, with or without communication bottlenecks.} We plot idealized wall-clock time (see \Cref{sec:wallclock_model}) for training by \dpfull and \dlc across compute nodes connected via high-, medium-, and low-bandwidth networks, for varying model sizes. For models up to 2.4B, we also vary global batch size. For 4B and 10B models, we use the batch size predicted by scaling laws, with discussion of fitting these scaling laws in \Cref{sec:scaling_laws}.
    \dlc is faster in almost all settings, due to its reduced communication and tolerance to larger batch sizes. Even for high-bandwidth networks, larger batch sizes reduce wall-clock time. We see similar results for $M \geq 4$ and for smaller models, but omit for visual clarity.}
    \label{fig:wallclock}
\end{figure}

\begin{tcolorbox}[colback=googlegreen, colframe=black, arc=4pt, boxsep=0.pt]%
\textbf{Finding 3: Optimal batch size.} \dlc increases the optimal batch size and moreover, the optimal global batch size increases with $M$ (see Figures \ref{fig:batch_size_eval_loss_reduced} and \ref{fig:batch_size_hellaswag_reduced}). This means that \dlc improves horizontal scalability relative to \dpfull (see \Cref{fig:wallclock}).
\end{tcolorbox}%

While DiLoCo with $M > 1$ often did slightly worse in terms of evaluation loss when picking the best experiment across all hyperparameters, it exhibited significantly improved performance with respect to batch size. While \dpfull and DiLoCo, $M = 1$ did well with small batch sizes, \dpfull's performance degrades quickly as batch size increases. By contrast, DiLoCo with any $M$ exhibits much more stable performance with respect to batch size. Examples of this phenomenon are in \Cref{fig:batch_size_eval_loss_reduced} for evaluation loss, and \Cref{fig:batch_size_hellaswag_reduced} for zero-shot accuracy on HellaSwag. As batch size increases, \dpfull becomes worse than DiLoCo with $M = 2, 4$, and eventually, $M = 8$.

One important consequence is that DiLoCo results in much more natural \emph{horizontal scalability}. Recall that in all cases, the token budget $D$ is a function only of $N$. This means that when using a batch size that is $4\times$ larger (for example), we do $4\times$ fewer training steps. For DiLoCo, this yields quite good performance, and can use more resources all at once, reducing total training time. By contrast, \dpfull seems to require much more serial training. This reduction in training time is compounded by reducing communication. To show these effects, we plot an idealized wall-clock time when training under networks of varying bandwidth in \Cref{fig:wallclock}. We see that \dlc's tolerance for larger batch sizes allows it to achieve comparable loss to \dpfull significantly faster, and that this is only amplified in low-bandwidth settings.

\begin{figure}[h!]
    \centering
    \includegraphics[width=0.5\linewidth]{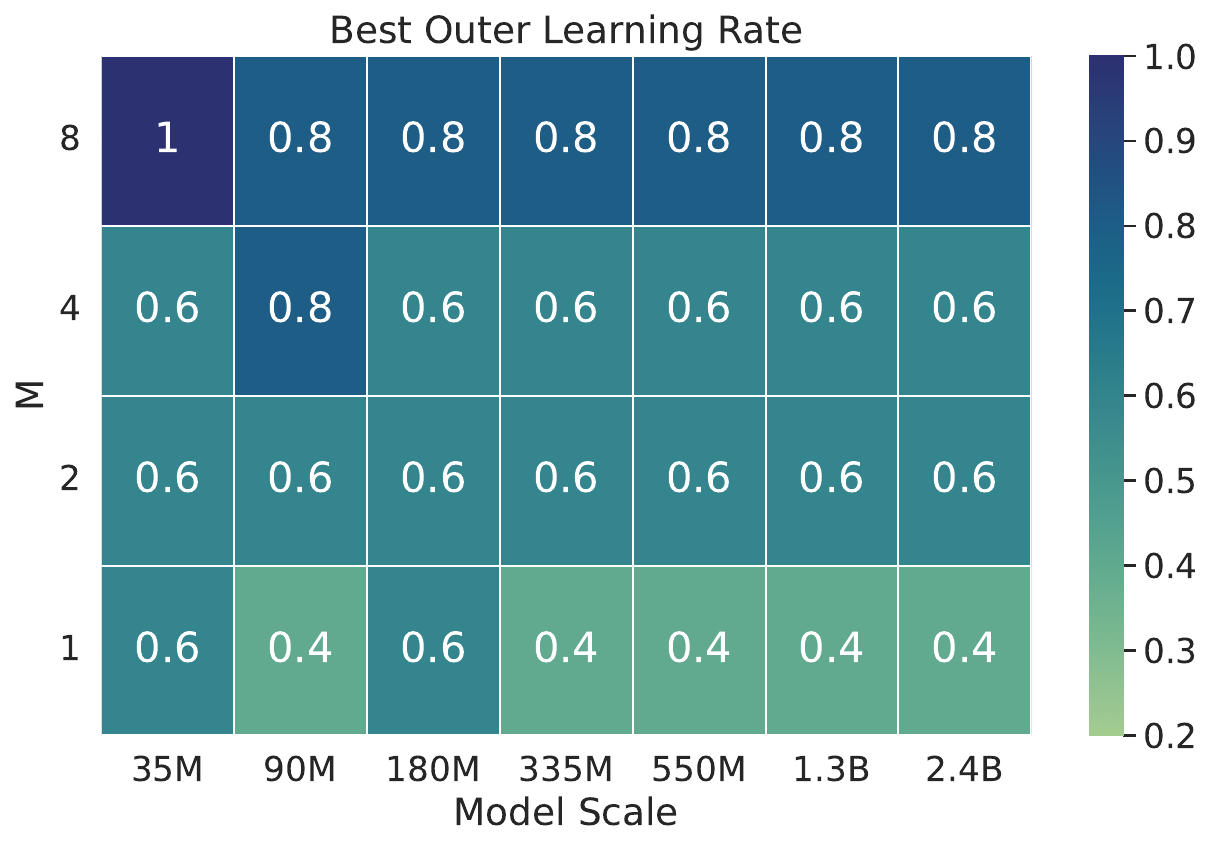}
    \vspace{-0.2cm}
    \caption{\textbf{Optimal outer learning rate is independent of model size.} We present the best outer learning rate for DiLoCo for varying $M$ and model sizes $N$. We select the best outer learning rate over $\{0.2, 0.4, 0.6, 0.8, 1.0\}$, optimizing over inner learning rate $\gamma$ and global batch size $B$. For sufficiently large models, the best outer learning rate is clearly constant.}
    \label{fig:best_olr}
\end{figure}

\begin{tcolorbox}[colback=googlegreen, colframe=black, arc=4pt, boxsep=0.pt]%
\textbf{Finding 4: Outer learning rate.} The optimal outer learning rate is essentially constant with respect to the model size $N$, but varies with $M$ (see \Cref{fig:best_olr}).
\end{tcolorbox}%

While optimal inner learning rate varies with model size $N$, the optimal outer learning rate $\eta$ for DiLoCo is independent of $N$ and depends only on $M$. As shown in \Cref{fig:best_olr}, for sufficiently large models ($N \geq$ 335M), the best $\eta$ for each $M$ is constant. Larger values of $M$ seem to necessitate larger $\eta$. This is consistent with prior findings that outer learning rate should increase as a function of number of clients in federated learning settings~\citep{charles2021large}.

\FloatBarrier

\section{Ablations}\label{sec:ablations}

\begin{figure}[ht]
\centering
\begin{subfigure}{0.4\linewidth}
    \centering
    \includegraphics[height=5cm]{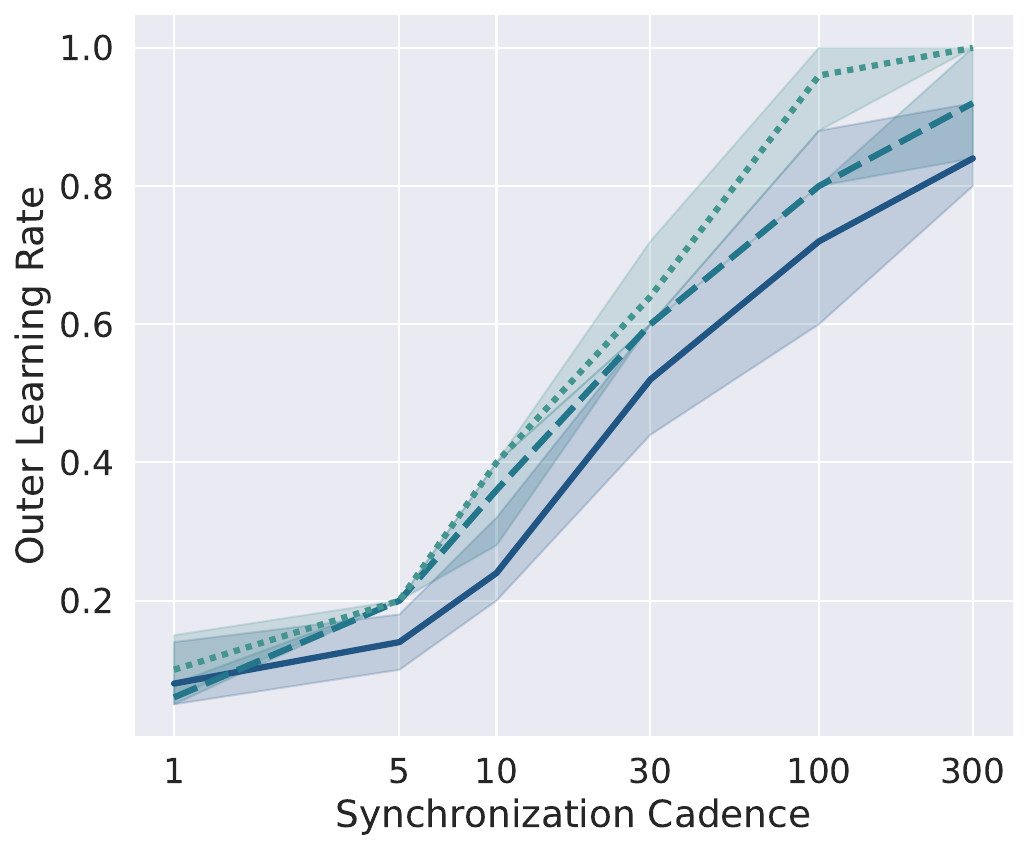}
    \caption{Optimal outer learning rate for each synchronization cadence. Shaded regions represent the variance across model sizes.}
\end{subfigure}
\hspace{1cm}
\begin{subfigure}{0.5\linewidth}
    \centering
    \includegraphics[height=5cm]{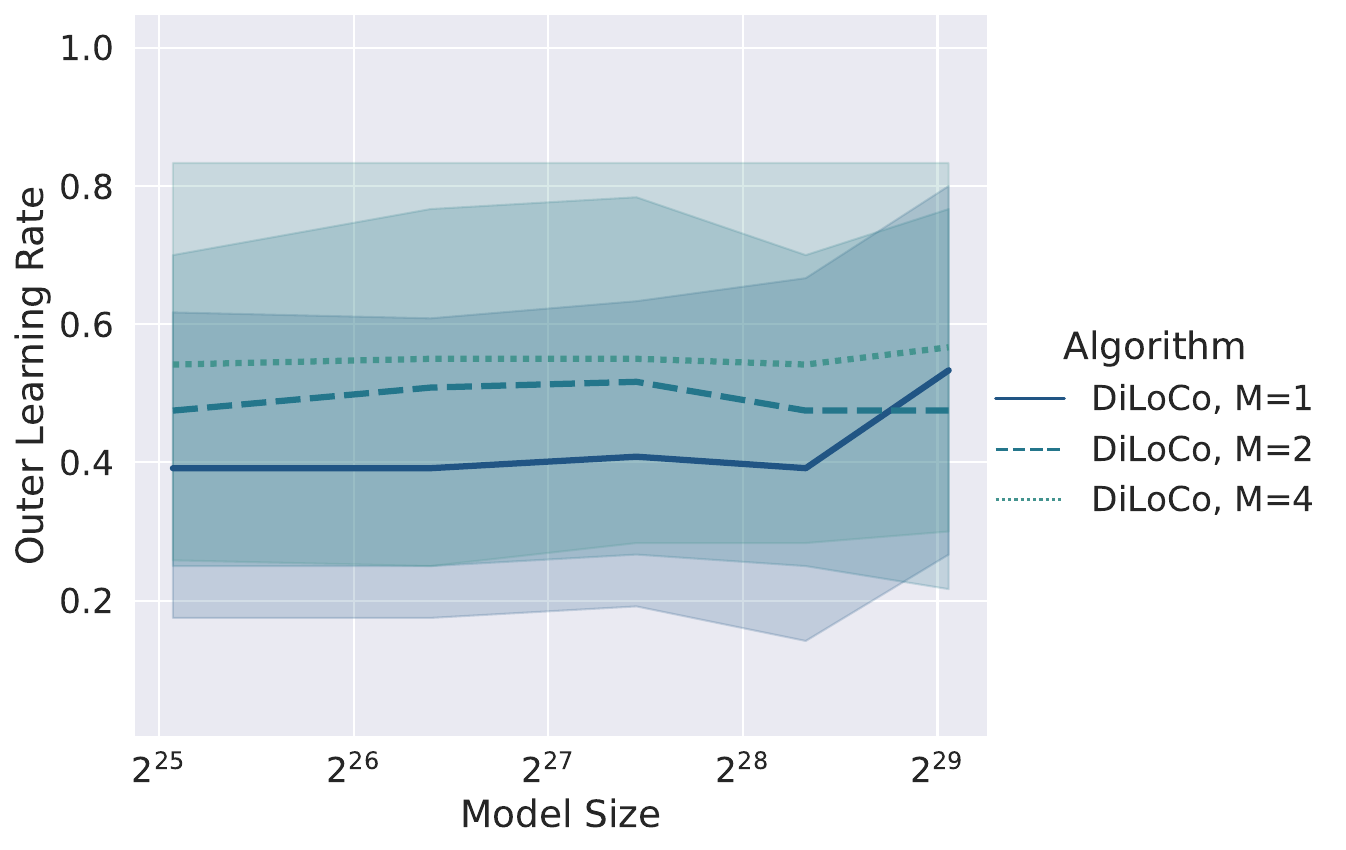}
    \caption{Optimal outer learning rate for each model size. Shaded regions represent the variance across synchronization cadences.}
\end{subfigure}
\caption{\textbf{Outer learning rate scales with $M$ and $H$, not $N$.} The optimal outer learning rate $\eta$ is a monotonically increasing function of synchronization cadence $H$ and $M$ (left), but essentially independent of model size $N$ (right). As in \Cref{fig:best_olr}, this means that we can tune outer learning rate at smaller model scales.}
\label{fig:olr_wrt_synch_cadence}
\end{figure}

\begin{figure}[ht]
    \centering
    \includegraphics[width=0.5\linewidth]{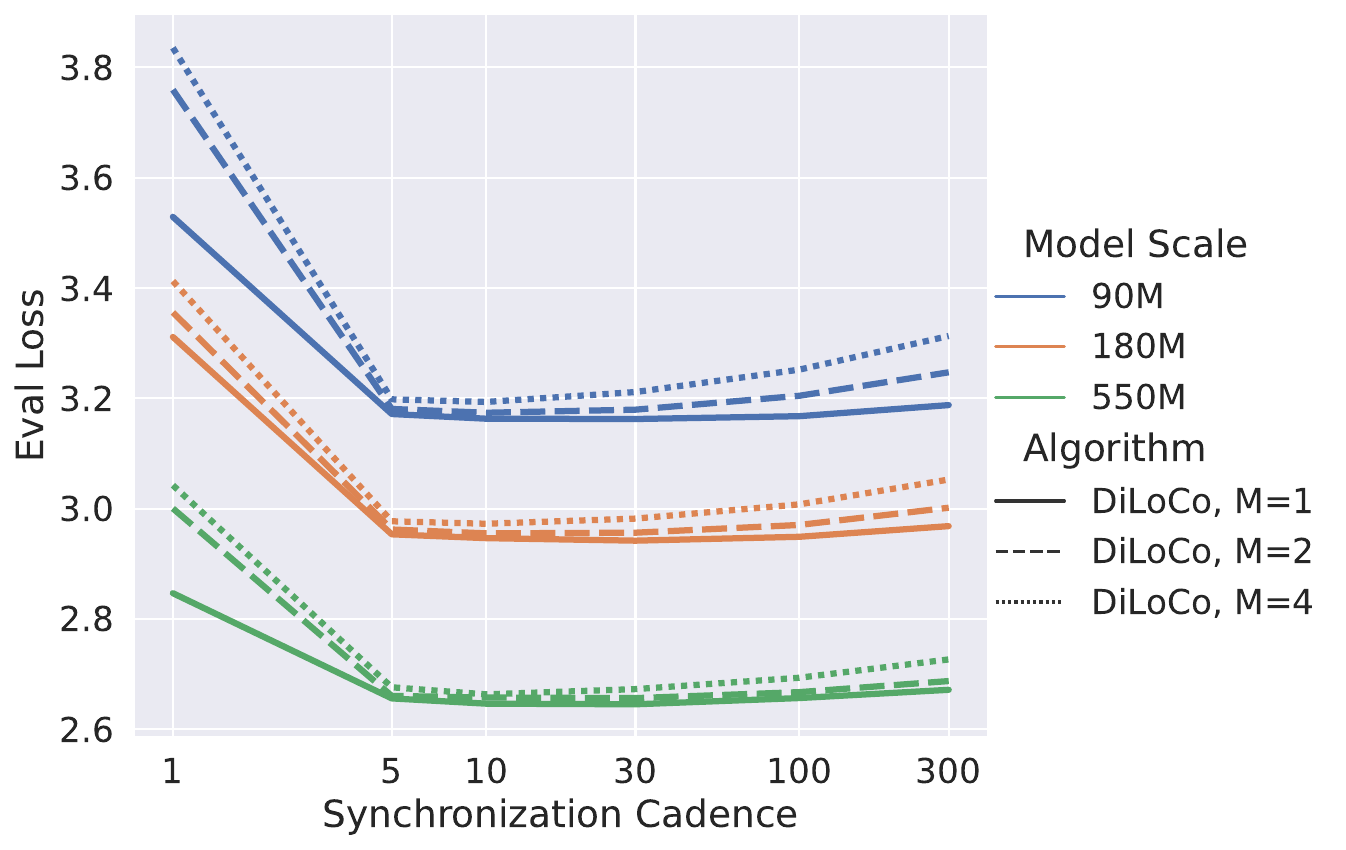}
    \caption{\textbf{Infrequent synchronization works better for larger models.} Outside of $H = 1$, which performs the worst, evaluation loss increases with $H$. However, the rate of increase is less pronounced for DiLoCo with $M = 1$ and for larger models, suggesting that large models can be synchronized quite infrequently.}
    \label{fig:loss_wrt_synch_cadence}
\end{figure}

\subsection{Synchronization Cadence}\label{sec:sync_cadence}

Our experiments above all used a synchronization cadence $H$ of 30 (ie. after every $H = 30$ inner optimization steps, DiLoCo performs an outer optimization step). We now study an ablation designed to ensure that our results are consistent with other values of $H$. To that end, we apply \dlc with varying $M$ and across various model sizes $N$. For each such setting, we take the optimal inner learning rate and global batch size from above, but perform a sweep over $H \in \{1, 5, 10, 30, 100, 300\}$ and $\eta \in \{0.05, 0.1, 0.2, 0.4, 0.6, 0.8, 1.0\}$. We first study whether the observation in \Cref{sec:exp_results}, that $\eta$ should be tuned independently of $N$, holds for other values of $H$.

We give results in the affirmative in \Cref{fig:olr_wrt_synch_cadence}. 
We see that across model scales, the optimal learning rate is essentially only a function of the number of replicas $M$ and the synchronization cadence $H$, and is essentially independent of model size $N$.
There is some slight variation, though this is likely due to not re-tuning the inner learning rate. Moreover, our results actually show a potentially surprising phenomenon: The optimal outer learning rate \emph{increases} with $H$. This is potentially counter-intuitive; as $H$ increases, the DiLoCo replicas may diverge more, and so one might expect that a more conservative learning rate is warranted. This is not the case.

Next, we analyze how $H$ impacts the evaluation loss of \dlc in \Cref{fig:loss_wrt_synch_cadence}. For all models, synchronizing every step ($H = 1$) performs the worst, but after this point all values of $H$ perform somewhat comparably. For a fixed $N$ and $M$, evaluation loss increases as $H$ increases. However, this increase is less pronounced for $M = 1$ and larger models. This yields an important finding: As the model size $N$ increases, we can actually perform synchronization across DiLoCo replicas less frequently, while nearly maintaining evaluation performance.

\paragraph{Compute utilization.}

\begin{figure*}[t]
\centering
\captionsetup[subfigure]{justification=centering}
\begin{subfigure}{0.325\linewidth}
  \centering
  \includegraphics[width=1\linewidth]{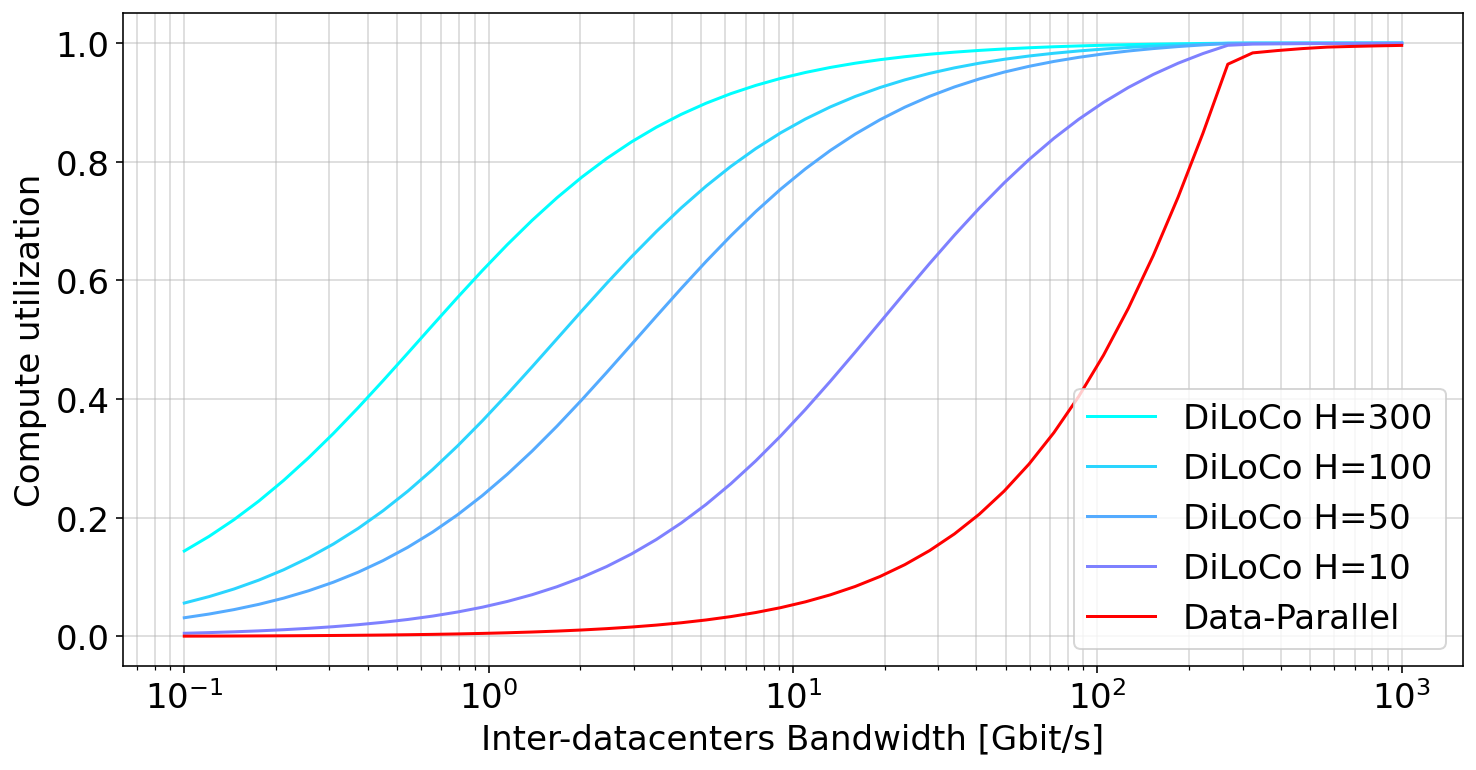}
  \caption{10B Chinchilla}
  \label{fig:bandwdith_chinchilla}
\end{subfigure}\hfill
\begin{subfigure}{0.325\linewidth}
  \centering
  \includegraphics[width=1\linewidth]{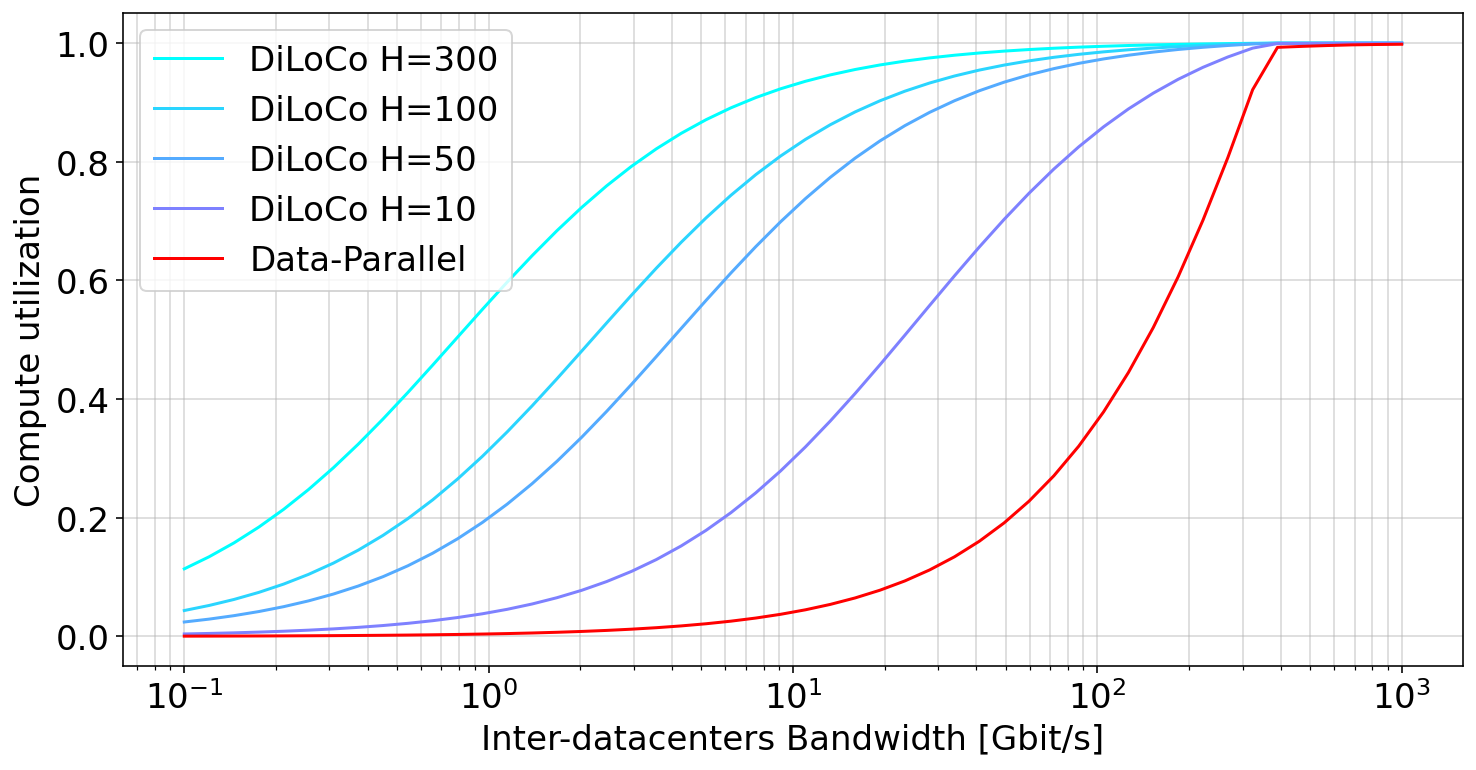}
  \caption{405B Llama3}
  \label{fig:bandwdith_llama}
\end{subfigure}\hfill
\begin{subfigure}{0.325\linewidth}
  \centering
  \includegraphics[width=1\linewidth]{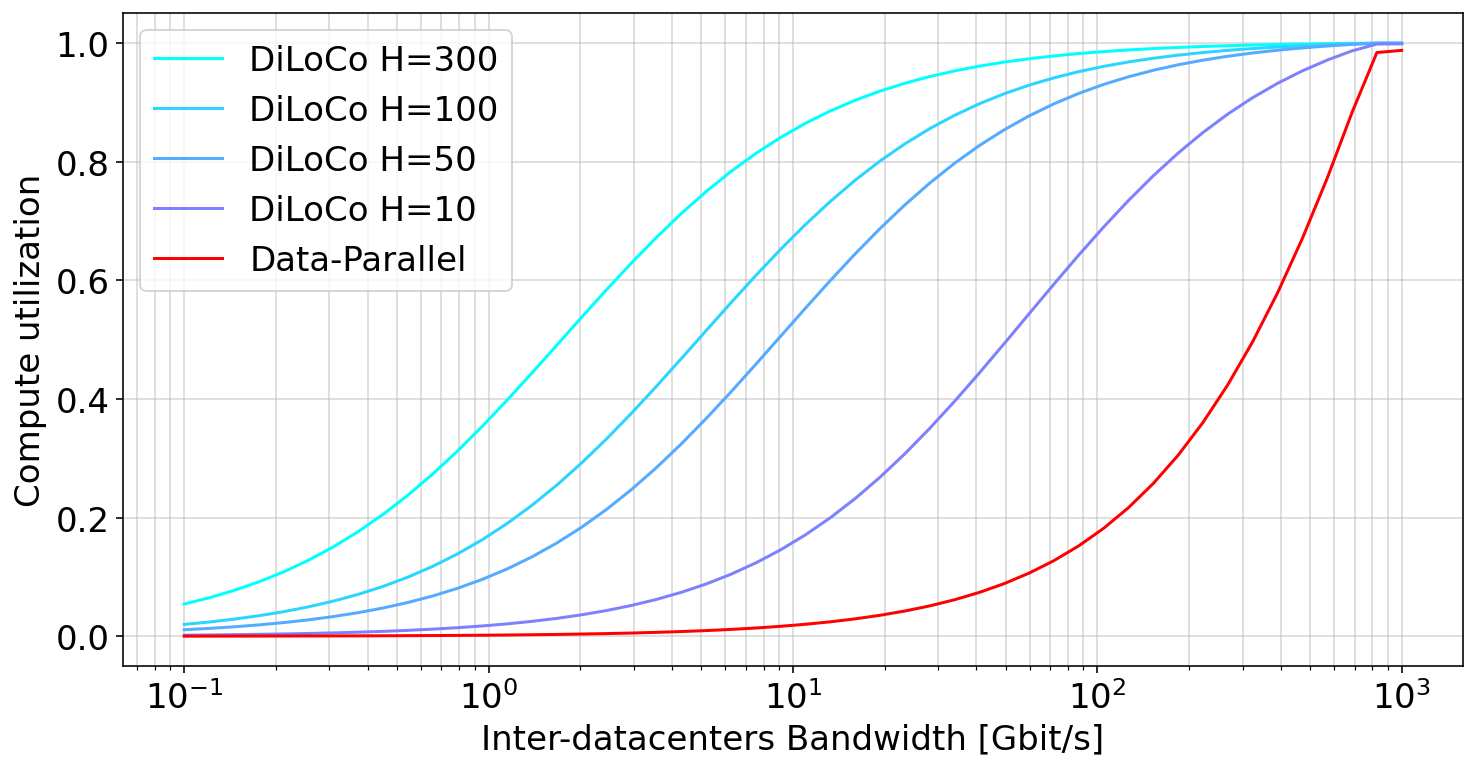}
  \caption{671B DeepSeek-V3}
  \label{fig:bandwdith_deepseek}
\end{subfigure}
\caption{\textbf{DiLoCo greatly increases compute utilization}. Here we present simulated compute utilization for \dlc and \dpfull across a range of bandwidth and synchronization cadences $H$. A compute utilization of 0.8 means 80\% of the time is spent in computation, and 20\% in communication. A higher value is better. We see similar results for other model sizes, but omit for visual clarity.}
\label{fig:bandwdith}
\end{figure*}

\begin{table*}[ht]
\small
\centering
\begin{tabular}{@{}ccc|l|ccccc@{}}
\toprule
\multirow{2}{*}{Architecture} & \multirow{2}{*}{Size} & \multirow{2}{*}{Step time} &   \multirow{2}{*}{Method} & \multicolumn{5}{c}{Gbit/s to reach a compute utilization $\texttt{CU} = $?} \\
&&&&$50\%$ &  $80\%$ &  $90\%$ &  $95\%$ &  $99\%$  \\
\midrule
\multirow{5}{*}{Chinchilla} & \multirow{5}{*}{10B} & \multirow{5}{*}{0.8s} & Data-Parallel & 104.8 & 184.2 & 222.3 & 222.3 & 390.7 \\
& & & DiLoCo, $H=1$ & 104.8 & 184.2 & 222.3 & 222.3 & 390.7 \\
& & & DiLoCo, $H=10$ & 16.0 & 49.4 & 86.8 & 152.6 & 222.3 \\
& & & DiLoCo, $H=50$ & \tabemphgood{3.0} & \tabemphgood{11.0} & 23.3 & 41.0 & 126.5 \\
& & & DiLoCo, $H=100$ & \tabemphgood{1.4} & \tabemphgood{6.2} & \tabemphgood{13.3} & 23.3 & 86.8 \\
& & & DiLoCo, $H=300$ & \tabemphbest{0.5} & \tabemphgood{2.0} & \tabemphgood{4.3} & \tabemphgood{9.1} & 41.0 \\
\midrule
\multirow{5}{*}{Llama3} & \multirow{5}{*}{405B} & \multirow{5}{*}{26s} & Data-Parallel & 126.5 & 222.3 & 268.3 & 323.8 & 323.8 \\
& & & DiLoCo, $H=1$ & 126.5 & 222.3 & 268.3 & 323.8 & 323.8 \\
& & & DiLoCo, $H=10$ & 19.3 & 72.0 & 126.5 & 184.2 & 268.3 \\
& & & DiLoCo, $H=50$ & \tabemphgood{3.6} & \tabemphgood{13.3} & 28.1 & 59.6 & 184.2 \\
& & & DiLoCo, $H=100$ & \tabemphgood{2.0} & \tabemphgood{7.5} & \tabemphgood{16.0} & 33.9 & 126.5 \\
& & & DiLoCo, $H=300$ & \tabemphbest{0.7} & \tabemphgood{3.0} & \tabemphgood{6.2} & \tabemphgood{13.3} & 59.6 \\ 
\midrule
\multirow{5}{*}{DeepSeek-V3} & \multirow{5}{*}{671B} & \multirow{5}{*}{20s} & Data-Parallel & 323.8 & 569.0 & 686.6 & 686.6 & 1000.0+ \\
& & & DiLoCo, $H=1$ & 323.8 & 569.0 & 686.6 & 686.6 & 1000.0+ \\
& & & DiLoCo, $H=10$ & 49.4 & 152.6 & 268.3 & 390.7 & 686.6 \\
& & & DiLoCo, $H=50$ & \tabemphgood{7.5} & \tabemphgood{33.9} & 72.0 & 126.5 & 390.7 \\
& & & DiLoCo, $H=100$ & \tabemphgood{4.3} & \tabemphgood{16.0} & \tabemphgood{41.0} & 72.0 & 268.3 \\
& & & DiLoCo, $H=300$ & \tabemphbest{1.7} & \tabemphgood{6.2} & \tabemphgood{13.3} & \tabemphgood{28.1} & 126.5 \\
\bottomrule
\end{tabular}
\caption{\textbf{Simulated compute utilization}. We estimate the step time based on the required flops using the rule proposed by \citet{kaplan2020scaling} and a max flop utilization of 60\%. We estimate the bandwidth (in Gbit/s) required to reach a level of compute utilization using \citep{douillard2025streaming}'s simulator. We highlight in light blue 10$\times$ reduction of bandwidth, and in dark blue 100$\times$ reduction.}
\label{tab:bandwidth}
\end{table*}

The synchronization cadence of \dlc is critical to training large-scale models distributed across the world. Indeed, less frequent synchronization (larger $H$) diminishes the bandwidth requirements of training. Following \cite{douillard2025streaming}, we simulate the amount of bandwidth required to have a compute utilization ($\frac{\text{compute time}}{\text{total time}}$) as large as possible for three types of LLMs: a 10B Chinchilla-style transformer~\citep{hoffmann2022training} in Fig.\ref{fig:bandwdith_chinchilla}, a 405B Llama3 model~\citep{dubey2024llama} in Fig.\ref{fig:bandwdith_llama}, and a 671B DeepSeek-v3 MoE~\citep{liu2024deepseek} in Fig.\ref{fig:bandwdith_deepseek}. We also report raw numbers in \Cref{tab:bandwidth}.

\FloatBarrier

\subsection{Overtraining}\label{sec:overtrain}

\begin{figure}[ht]
    \centering
    \includegraphics[width=0.7\linewidth]{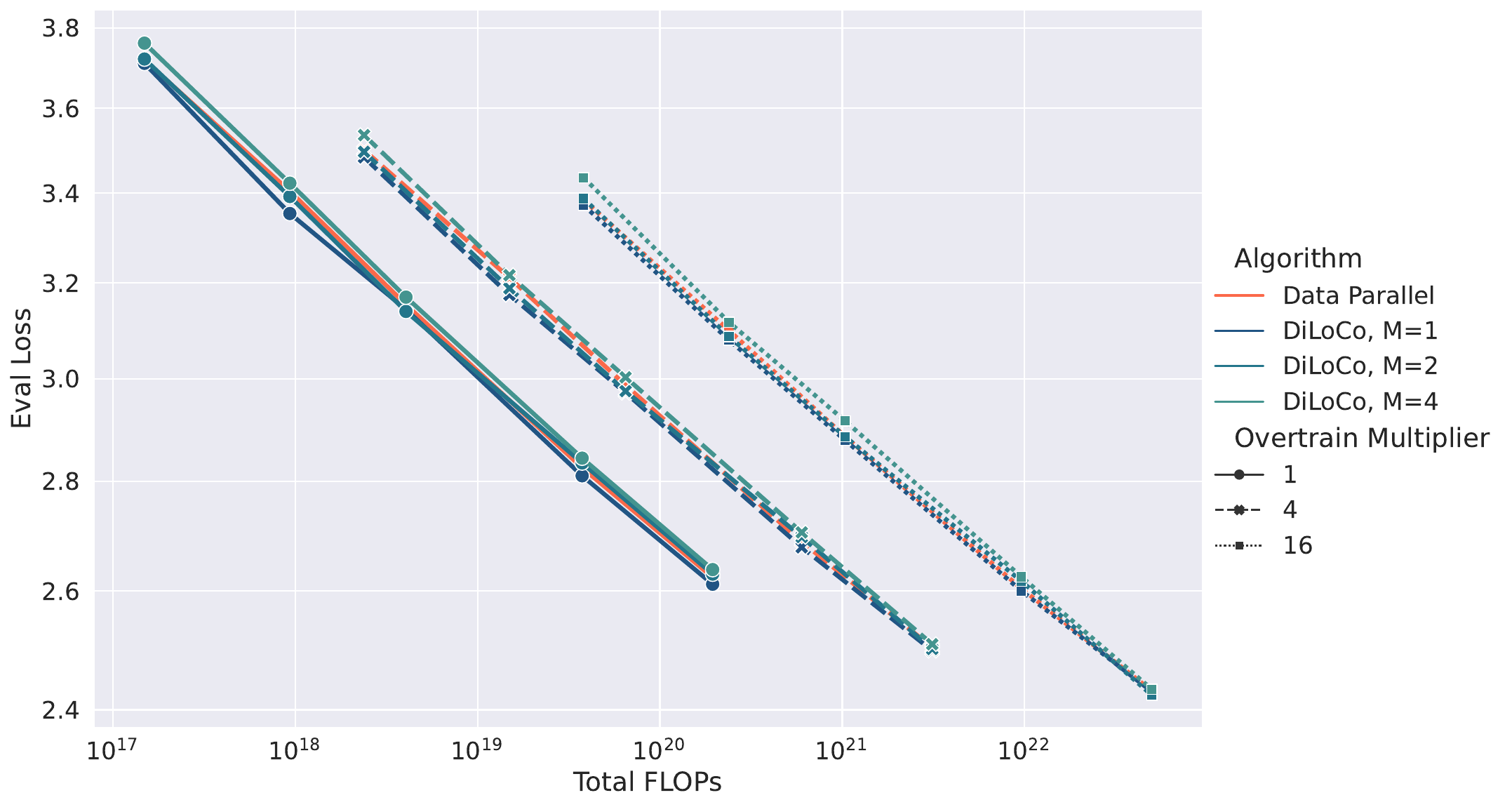}
    \caption{\textbf{DiLoCo scales reliably with overtraining.} For each overtrain multiplier, the curves of loss with respect to FLOPs are all essentially parallel lines in log-space. We see that variations in loss from changing the algorithm is dominated by changing the model size or amount of overtraining. Moreover, \dlc with $M = 1$ is slightly better than all other algorithms, including \dpfull, at all overtraining amounts and model sizes.}
    \label{fig:overtrain_results}
\end{figure}

In all of the above experiments, we use the Chinchilla-optimal amount of tokens for each model size~\citep{hoffmann2022training}. However, in many settings (e.g. when training a model with inference costs in mind) it is beneficial to perform \emph{overtraining}, where we use more tokens~\citep{gadre2024language}. We want to make sure that our results are robust to the amount of overtraining. To that end, we perform a swath of ablation studies on various \emph{overtraining multipliers}. Given an overtraining multiplier $\lambda \geq 1$, we train on $D = 20N\lambda$ tokens, so that $\lambda = 1$ corresponds to the Chinchilla-optimal number of tokens.

For our experiments, for varying model sizes $N$ and algorithms, we take the best-performing hyperparameters from our results above (ie. (global) batch size $B$, (inner) learning rate $\gamma$, and outer learning $\eta$ when using DiLoCo). We then train for $D = 20 N\lambda $ tokens on the Dolma dataset~\citep{soldaini2024dolma}, as it has more tokens than C4. Since we use QK-layernorm, we generally avoid the need to re-tune learning rates~\citep{gadre2024language}. We take the resulting models and evaluate them on the same evaluation set as before, the validation split of C4. We vary the overtrain multiplier $\lambda$ over $\{1, 4, 16\}$. Note that retraining for $\lambda = 1$ is necessary since we changed the underlying dataset. We train with \dpfull as well as DiLoCo with $M \in \{1, 2, 4\}$.

Our results are presented in \Cref{fig:overtrain_results}. We see that qualitatively, the scaling remains essentially unchanged as we do more overtraining. We emphasize that we did not re-tune any hyperparameter in these experiments. For each model size and algorithm, we simply took the best-performing hyperparameters from our Chinchilla-optimal experiments on C4. This means that the consistency of DiLoCo as we overtrain held despite the fact that for $M > 1$, we used much larger batch sizes than \dpfull.

\begin{figure}[thb]
    \centering
    \begin{subfigure}[b]{0.3\linewidth}
        \includegraphics[height=5cm]{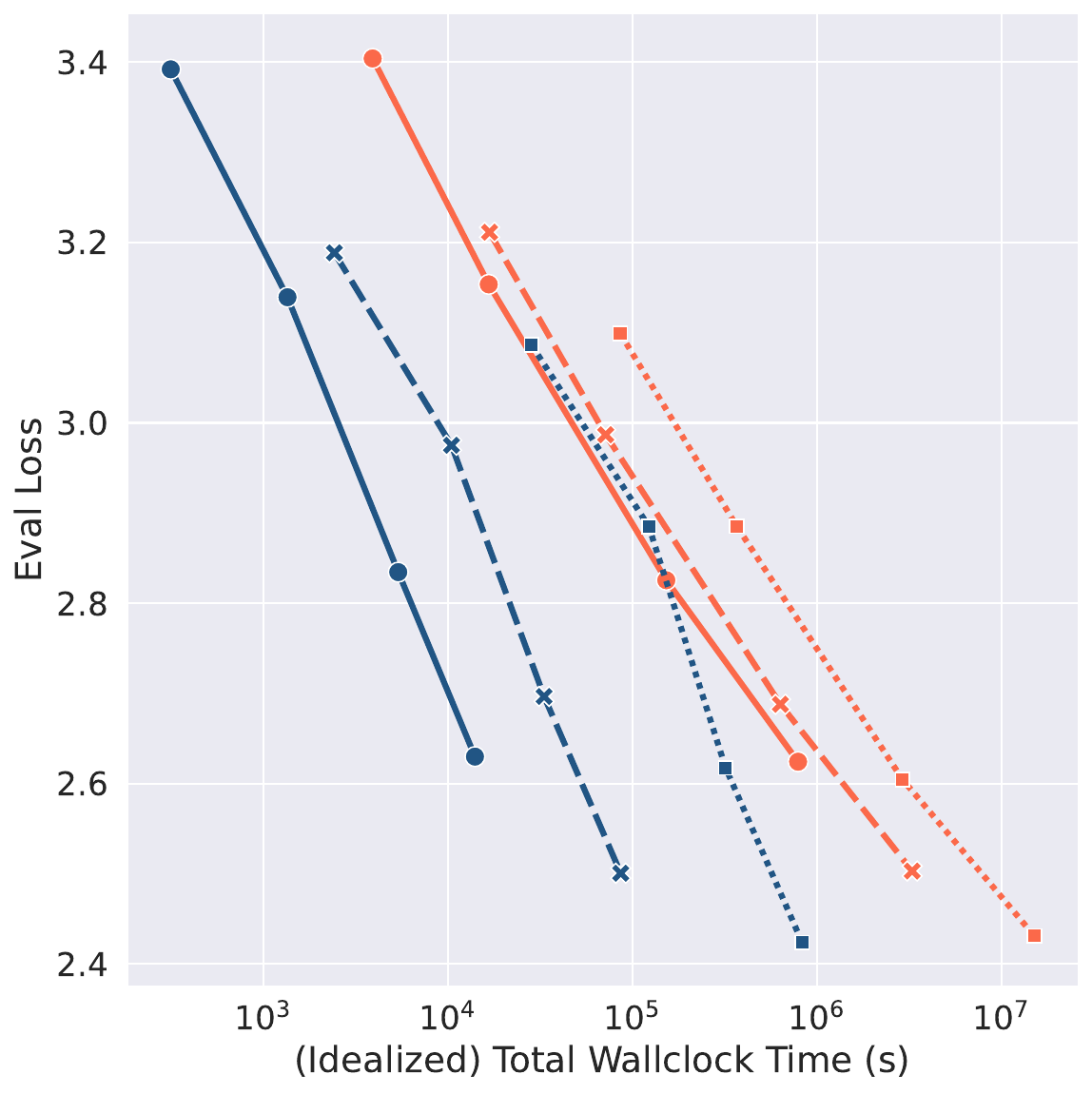}
        \caption{Network with a bandwidth of 10 gigabits/s and a latency of $10^{-2}$ seconds (\textbf{low-bandwidth}).}
    \end{subfigure}
    \hfill
    \begin{subfigure}[b]{0.3\linewidth}
        \includegraphics[height=5cm]{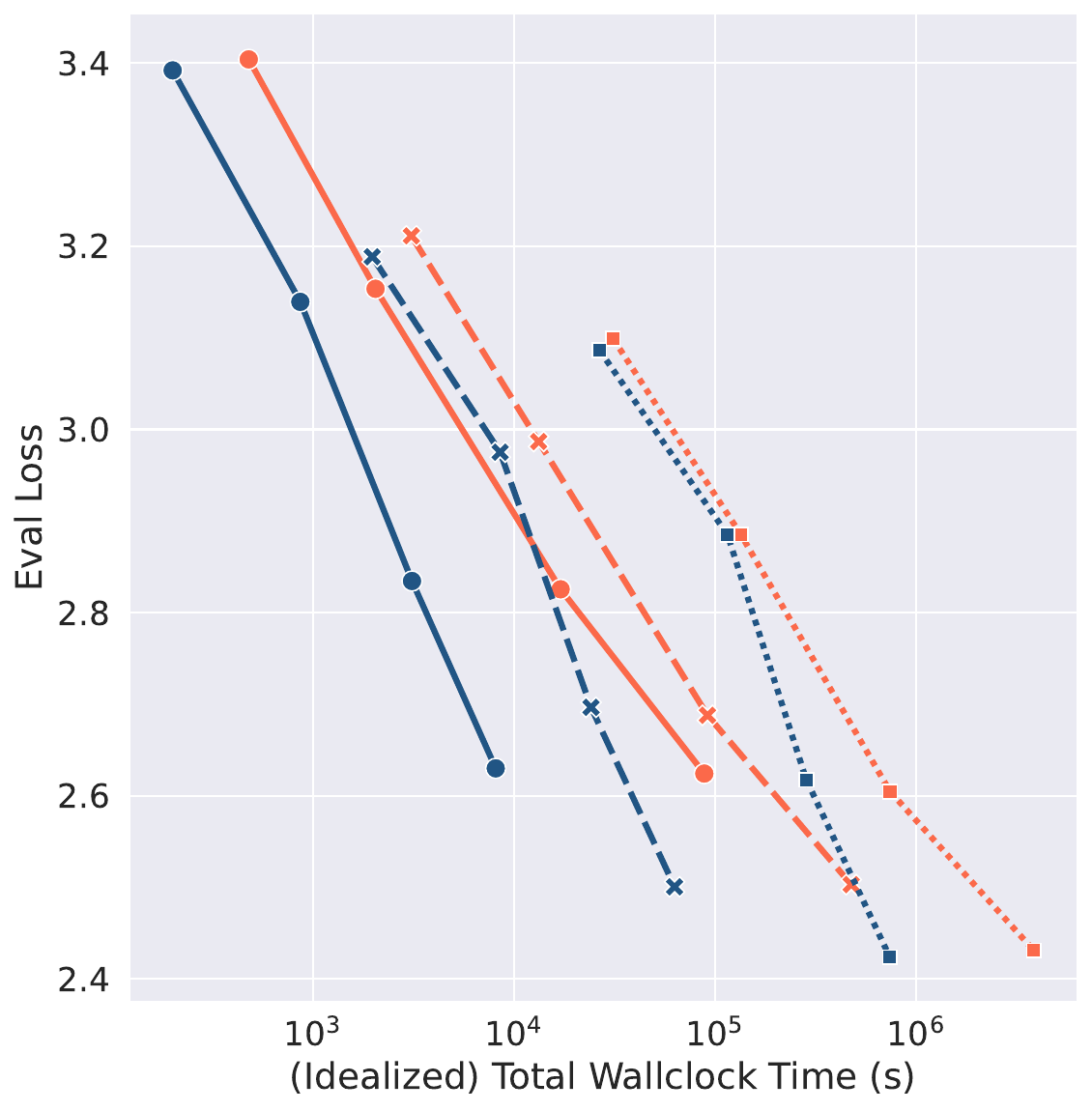}
        \caption{Network with a bandwidth of 100 gigabits/s and a latency of $10^{-3}$ seconds (\textbf{medium-bandwidth}).}
    \end{subfigure}
    \hfill
    \begin{subfigure}[b]{0.35\linewidth}
        \includegraphics[height=5cm]{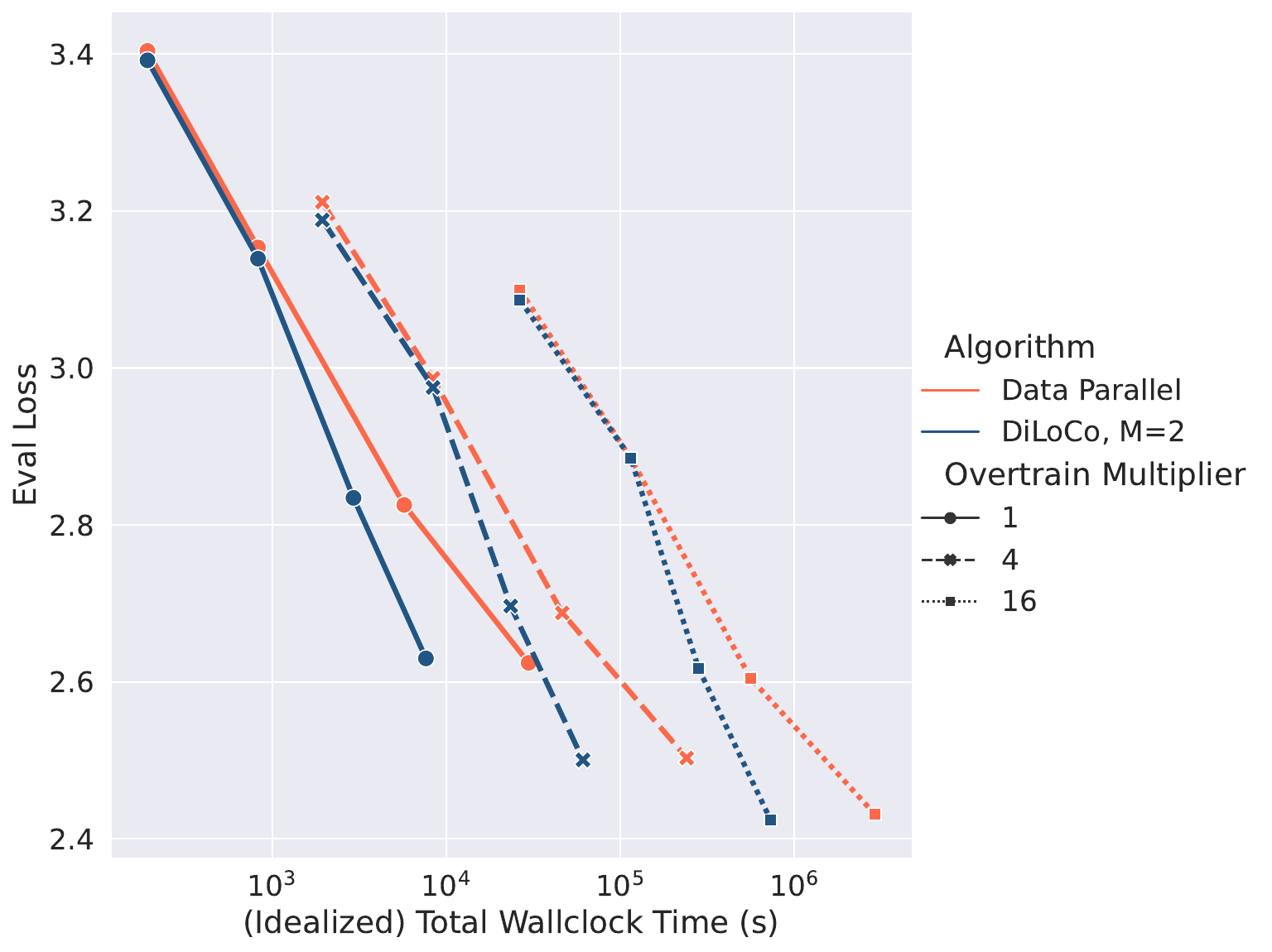}
        \caption{Network with a bandwidth of 400 gigabits/s and a latency of $10^{-4}$ seconds (\textbf{high-bandwidth}).}
    \end{subfigure}
    \caption{\textbf{\dlc speeds up overtraining by leveraging horizontal scalability.} For \dpfull and \dlc, $M = 2$, we plot idealized wall-clock time (see \Cref{sec:wallclock_model}) for training by \dpfull and \dlc, $M=2$ across compute nodes connected via high-, medium-, and low-bandwidth networks. For each algorithm and overtraining amount, we display lines represent varying model sizes, from 335M parameters to 2.4B.
    \dlc is faster in all settings, due to both its reduced communication and its tolerance to larger batch sizes. Even in the high-bandwidth setting, the larger batch sizes increase horizontal scalability, reducing end-to-end wallclock time.
    We see similar results for $M \geq 4$ and for smaller models, but omit for visual clarity.
    }
    \label{fig:ot_wallclock}
\end{figure}

To illustrate this, we plot idealized training time of \dpfull and \dlc with $M = 2$ for different overtraining amounts in \Cref{fig:ot_wallclock}. \dlc speeds up overtraining by reducing communication costs, and utilizing larger batch sizes, therefore requiring fewer serial training steps. This suggests that \dlc is a significant boon in overtraining, as we can amortize compute time (which can be quite long for overtraining) via horizontal scalability.

\FloatBarrier

\section{Scaling Laws}\label{sec:scaling_laws}

We now discuss the process we used to fit scaling laws to our empirical results. Recall that for each algorithm (\dpfull or \dlc with $M \in \{1, 2, 4, 8\}$) we ran extensive hyperparameter sweeps on models of size $N$ up to 2.4B. To fit a scaling law for loss $L$ as a function of $N$, we pick the best hyperparameters (in terms of evaluation loss) for each $N$, aggregate the loss values, and fit some kind of function for $L(N)$, such as a power law~\citep{kaplan2020scaling}. We will also fit scaling laws to the optimal hyperparameters.

We will fit two types of scaling laws for DiLoCo: independent fits for each value of $M$, and joint fits where we fit a single scaling law as a function of $N$ and $M$ simultaneously.

\subsection{Independent scaling laws}

\paragraph{Scaling laws for loss.} We first fit scaling laws for the loss obtained by Data-Parallel training. We fit a power law to the evaluation loss of \dpfull as a function of $N$ via the power law approximation $L(N) \approx AN^{\alpha}$. Note that this can easily be done via applying linear fit techniques to $\log(L)$, and is not sensitive to things like initial values of $A, \alpha$. The resulting power law is in the first row of \Cref{table:loss_ind_fit}.

We mirror this above for \dlc when doing independent fits. For each value of $N, M$, we record the lowest loss value across all hyperparameters. We can then fit power law $L_M(N) : = L(N, M) \approx AN^{\alpha}$ for each $M$. The results are given in \Cref{table:loss_ind_fit}.

\begin{table}[htb]
\centering
    \caption{Power law approximations for loss $L(N) \approx AN^{\alpha}$.}
    \label{table:loss_ind_fit}
    \begin{tabular}{ccc}
        \toprule
         & $A$ & $\alpha$ \\
        \midrule
        \dpfull & $18.129$ & $-0.0953$ \\
        \dlc, $M = 1$ & $18.363$ & $-0.0961$ \\
        \dlc, $M = 2$ & $18.768$ & $-0.0969$ \\
        \dlc, $M = 4$ & $19.762$ & $-0.0992$ \\
        \dlc, $M = 8$ & $21.051$ & $-0.1018$ \\
        \bottomrule
    \end{tabular}
\end{table}

The results show that \dpfull and \dlc see similar predicted reductions in loss as a function of $N$. Notably, the fit parameters suggest that DiLoCo, $M = 1$ outperforms \dpfull at essentially all but the absolute smallest model scales. This mirrors the results discussed in \Cref{sec:exp_results}.

\paragraph{Scaling laws for hyperparameters.} For \dpfull, we fit scaling laws for learning rate $\gamma$ and batch size $B$. For \dlc, we fit scaling laws for inner learning rate $\gamma$ and global batch size $B$. Given their analogous role in the algorithms, we fit them in the same way. For (inner) learning rate, we use the same approach as fitting scaling laws for loss: for each $N$ (and $M$, for \dlc), we select the best hyperparameters, and fit a power law. The results are in \Cref{table:lr_ind_fit}.

\begin{table}[t]
    \centering
    \begin{minipage}{0.45\linewidth}
        \centering
        \caption{Power law approximations for \\ (inner) learning rate $\gamma(N) \approx AN^{\alpha}$.}
        \label{table:lr_ind_fit}
        \begin{tabular}{ccc}
            \toprule
             & $A$ & $\alpha$ \\
            \midrule
            \dpfull & $16319.2$ & $-0.819$ \\
            \dlc, $M = 1$ & $74620.6$ & $-0.945$ \\
            \dlc, $M = 2$ & $3978.82$ & $-0.780$ \\
            \dlc, $M = 4$ & $4512.99$ & $-0.789$ \\
            \dlc, $M = 8$ & $618986$ & $-1.102$ \\
            \bottomrule
        \end{tabular}
    \end{minipage}
    \hspace{0.1cm}
    \begin{minipage}{0.45\linewidth}
        \centering
        \caption{Power law approximations for \\ (global) batch size $B(N) \approx AN^{\alpha}$.}
        \label{table:bsz_ind_fit}
        \begin{tabular}{ccc}
            \toprule
             & $A$ & $\alpha$ \\
            \midrule
            \dpfull &  $0.22592$ & $0.281$ \\
            \dlc, $M = 1$ & $0.01361$ & $0.435$ \\
            \dlc, $M = 2$ & $0.00769$ & $0.479$ \\
            \dlc, $M = 4$, & $0.00535$ & $0.510$ \\
            \dlc, $M = 8$ & $0.01859$ & $0.455$ \\
            \bottomrule
        \end{tabular}
    \end{minipage}
\end{table}

For (global) batch size, we alter this slightly. As discussed in \Cref{sec:exp_setup}, our sweeps use powers of $2$ for batch size, in order to saturate compute. However, the optimal batch size may be between these values. To account for this, we first fit a quadratic approximation to the batch size. Specifically, for each value of $N$ we look at the loss as a function of $\log_2(B)$ (when using the best learning rate for that $B$), and fit a quadratic to this function. We select the minima of those quadratics and fit a power law to them, as a function of $N$. The results are in \Cref{table:bsz_ind_fit}.

DiLoCo has a third hyperparameter we could fit scaling laws to: the outer learning rate. However, as shown in \Cref{sec:exp_results}, the optimal outer learning rate is (for sufficiently large models) seemingly constant. Therefore, a scaling law would seemingly not yield any improved predictive performance over simply using the best outer learning rate for each $M$ (see \Cref{fig:best_olr}).

\subsection{Joint scaling laws}

\begin{table}[htb]
    \centering
    \caption{Joint power law approximations $f(N, M) = AN^{\alpha}M^{\beta}$ for the loss $L$, inner learning rate $\gamma$, and batch size $B$ of DiLoCo.}
    \label{table:joint_fit}
    \begin{tabular}{cccc}
        \toprule
        & $A$ & $\alpha$ & $\beta$ \\
        \midrule
        $L$ & $19.226$ & $-0.0985$ & $0.0116$ \\
        $\gamma$ & $22256$ & $-0.8827$ & $0.2929$ \\
        $B$ & $0.00709$ & $0.4695$ & $0.3399$\\
        \bottomrule
    \end{tabular}
\end{table}

Alternatively, we can fit joint power laws to various facets of DiLoCo, using a two-variable power law $f(N, M) \approx AN^{\alpha}M^{\beta}$. We do this for loss $L$, inner learning rate $\gamma$ and global batch size $B$. For the first two, we select, for each value of $N, M$, the best learning rate and loss. For batch size we do the same, but using the quadratic approximations from the section above. We can then fit a joint power law via standard linear regression techniques. The resulting power laws are in \Cref{table:joint_fit}. Just as with the independent fits, we do not attempt to model outer learning, as the optimal value is independent of $N$.

\subsection{Measuring goodness-of-fit}

Now that we have two different ways of developing scaling laws for DiLoCo, we can attempt to ask which one yields better predictions. First we do this via leave-one-out validation. Specifically, we use the same methodology as above to fit scaling laws for $L$, $\gamma$, and $B$, but only using data up to $N =$ 1.3B parameters, leaving out our data on $N$ = 2.4B parameters. We then use the scaling law to predict the optimal value for $L, \gamma$, and $B$ at $N$ = 2.4B parameters, across different values of $M$.

\begin{table}[htb]
    \centering
    \caption{\textbf{Joint fit scaling laws match or beat independent fit.} Here we give the residuals for scaling law predictions for $N$ = 2.4B and varying $M$. We compare the residual of independent and joint fitting strategies in predicting loss $L$, inner learning rate $\gamma$, and global batch size $B$. For the average residuals, we highlight which of independent or joint achieved a lower residual. We see that the joint fit matches independent for $L$ and $B$, but does better at predicting $\gamma$.}
    \label{table:leave_out_2p5b}
    \begin{tabular}{cc|c|c|c}
        \toprule
        & Fit & $L$ & $\gamma$ & $B$ \\
        \midrule
        \multirow{2}{*}{$M = 1$}
        & Independent & \tabemph{0.011} & 0.35 & \tabemph{0.00088} \\
        & Joint & 0.019 & \tabemph{0.14} & 0.19 \\
        \midrule
        \multirow{2}{*}{$M = 2$}
        & Independent & \tabemph{0.0099} & \tabemph{0.18} & 0.44 \\
        & Joint & 0.013 & 0.29 & \tabemph{0.28} \\
        \midrule
        \multirow{2}{*}{$M = 4$}
        & Independent & 0.012 & \tabemph{0.051} & 0.25 \\
        & Joint & \tabemph{0.0082} & 0.086 & \tabemph{0.11} \\
        \midrule
        \multirow{2}{*}{$M = 8$}
        & Independent & 0.014 & 0.62 & \tabemph{0.076} \\
        & Joint & \tabemph{0.0076} & \tabemph{0.23} & 0.19 \\
        \midrule
        \multirow{2}{*}{Average over $M$}
        & Independent & \tabemphgood{0.012} & 0.30 & \tabemphgood{0.19} \\
        & Joint & \tabemphgood{0.012} & \tabemphgood{0.19} & \tabemphgood{0.19}\\
        \bottomrule
    \end{tabular}
\end{table}

Given a predicted value $\tilde{y}$ and an actual optimal value of $y$, we compute the \emph{residual} of our prediction as the mean absolute error of the logarithm: $\text{res}(y, \tilde{y}) = \lvert\log(y) - \log(\tilde{y})\rvert$. 
We use this measure as it works well for all three variables simultaneously, despite the fact that they vary greatly in their scale. For each $M \in \{1, 2, 4, 8\}$, we compute the predicted value of the three parameters above, and measure the residual relative the actual optimal value at $N$ = 2.4B. We report these values, as well as their average across $M$, in \Cref{table:leave_out_2p5b}.

Our results generally show that both approach is generally valid (as there is no clear winner) but also that there is significant variation in residuals between $M$. That being said, we see that on average, while the individual fit is slightly better at predicting the loss and global batch size, the independent fit is significantly better at predicting inner learning rate.

\subsection{Extrapolating to larger models}

We use the independent and joint fits to predict optimal hyperparameters for \dpfull and \dlc with $M \in \{1, 2, 4\}$ at 4B and 10B model scales. Note that for \dpfull, we can only use independent fits. We run training on these models with these hyperparameters, using a Chinchilla-optimal token budget, and compare the results.

\begin{table}[htb]
    \centering
    \caption{\textbf{Joint fit hyperparameters extrapolate well to larger models.} Here we show the evaluation results on 4B and 10B models, using hyperparameters predicted by individual and joint scaling laws. We highlight \dlc evaluation results that were better (ie. lower) than \dpfull. We see that while \dlc with $M = 2$ does better than \dpfull with either independent or joint fit rules, \dlc $M = 1$ only does better when using joint fit.}
    \begin{tabular}{cccc}
    \toprule
    \multirow{2}{*}{Algorithm} & \multirow{2}{*}{Fit Method} & \multicolumn{2}{c}{Loss} \\
    \cmidrule{3-4}
    & & 4B & 10B \\
    \midrule
    \dpfull & Independent & 2.224 & 2.090 \\
    \hline
    \multirow{2}{*}{\dlc, $M = 1$}
        & Independent & 2.229 & 2.103 \\
        & Joint & \tabemphgood{2.219} & \tabemphgood{2.086} \\
    \hline
    \multirow{2}{*}{\dlc, $M = 2$}
        & Independent & \tabemphgood{2.218} & \tabemphgood{2.083} \\
        & Joint & \tabemphgood{2.220} & \tabemphgood{2.086} \\
    \hline
    \multirow{2}{*}{\dlc, $M = 4$}
        & Independent & 2.232 & 2.098 \\
        & Joint & 2.230 & 2.096 \\
    \bottomrule
    \end{tabular}
\end{table}

We see two important facets. First, unlike results above, at 4B and 10B scales we see that \dlc with $M = 2$ actually outperforms both \dpfull and \dlc, $M = 1$, regardless of using individual or joint fit approaches. Second, we see that \dlc, $M = 1$ requires the joint fit to do better than \dpfull. Other than in this case, joint and independent fits perform comparably throughout. All in all, the joint fit approach to hyperparameters appears to have a slight edge over individual fit in extrapolating. Combined with its ability to also extrapolate to larger $M$, we generally recommend the joint fit approach for all hyperparameters.

\begin{figure}[htb]
    \centering
    \includegraphics[width=\linewidth]{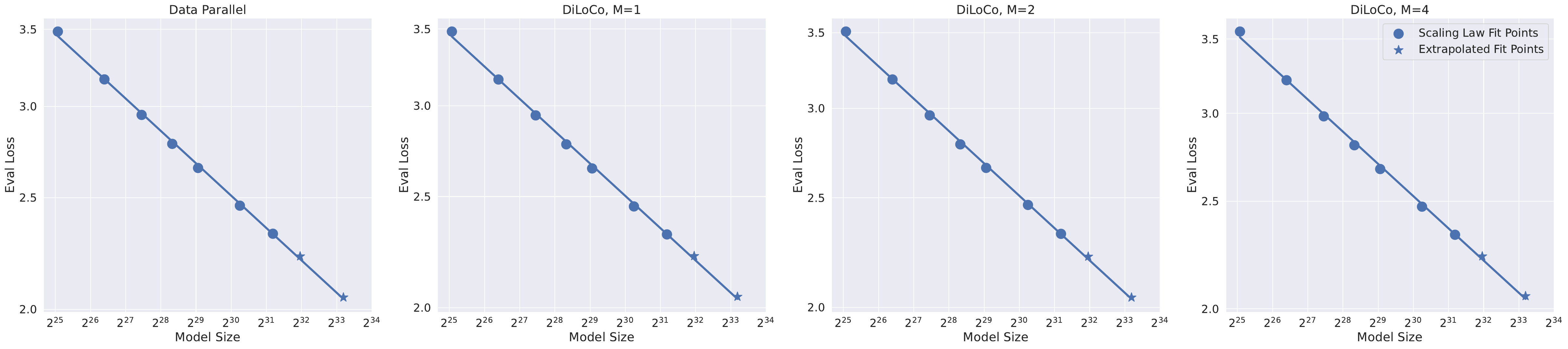}
    \caption{\textbf{DiLoCo scaling laws extrapolate well to larger models.} We present loss scaling laws for \dpfull and \dlc. Pictured are both the loss values to form the scaling law (by training models up to a scale of 2.4B) and loss values attained on larger models (4B and 10B). While we present the individual-fit scaling laws for simplicity, the joint fit also predicts loss well.}
    \label{fig:extrapolated_scaling}
\end{figure}

We now use these loss values to see how they compare to the scaling laws fit above. We generally find that the loss values are predicted very well, within a few percentage points of the loss predicted by the scaling laws. We present the fit scaling law and extrapolated loss values in \Cref{fig:extrapolated_scaling}.

\subsection{Parametric Function Fitting} While power laws have useful properties, various works on scaling laws have often found it useful to fit more complex functions to the data~\citep{hoffmann2022training}. In particular, a power law such as $AN^{\alpha}$ will only ever tend towards $0, 1, $ or diverge to $\infty$ as $N \to \infty$. In many cases, it makes sense to fit more complex functions.

We are particularly interested in parametric forms for our joint scaling laws. While \citet{hoffmann2022training} use a risk decomposition argument to decompose loss as a function of $N$ and $D$, it is not immediately clear how to do decompose $L(N, M)$. To that end, we use an empirical approach. We develop candidate functions, and determine which does best in an extrapolative sense. We use the following functional forms:
\begin{enumerate}
    \item $L(N, M) \approx AN^{\alpha}M^{\beta}$
    \item $L(N, M) \approx AN^{\alpha}M^{\beta} + C$
    \item $L(N, M) \approx AN^{\alpha + \beta M} + C$
    \item $L(N, M) \approx AN^{\alpha} + BM^{\beta} + C$
\end{enumerate}

The first is included for comparison's sake, as it recovers the power law scaling law used above. Fitting more sophisticated functions can be much more sensitive to initial values of parameters, and also sensitive to outlier data. Therefore, when fitting these functions to our loss values above, we use the general strategy proposed by \citet{hoffmann2022training}. 

In detail, let $f_Q(N, M)$ denote one of the functional forms above, where $Q$ represents the set of parameters to be fit (e.g. $Q = \{A, \alpha, \beta\}$ for the first). Let Huber$_\delta$ denote the Huber loss with parameter $\delta$. Let $\mathcal{N}, \mathcal{M}$ denote the set of values of $N$ and $M$ considered. For each $N, M$, we have an empirical loss $L(N, M)$, and some estimate of the loss $f_Q(N, M)$. We then solve the following minimization problem:
\[
\min_{Q} \sum_{N \in \mathcal{N}} \sum_{M \in \mathcal{M}} \text{Huber}_\delta\bigg(\log f_Q(N, M) - \log L(N, M)\bigg).
\]
We minimize this via L-BFGS, using some initialization $Q_0$ for the parameters. We repeat this process for 256 random initializations $Q_0$ of the parameters. We hold out all loss values at the $N$ = 2.4B scale, and select the parameters $Q$ that best fit the held-out data, measured in terms of the average residual $\lvert\log f_Q(N, M) - \log L(N, m)\rvert$ over all $M$.

\begin{table}[htb]
    \centering
    \caption{\textbf{Parametric function fitting improves joint scaling laws.} We showcase various parametric approximations to the empirical loss function $L(N, M)$, along with their validation error on held-out loss data at the $N$ = 2.4B scale. We see that joint power laws (the first) and classical loss decomposition (the last) are significantly worse at predicting loss on the held-out data.}
    \label{tab:parametric_joint_laws}
    \begin{tabular}{lc}
        \toprule
        Parametric form & Average Residual \\
        \midrule
        $L(N, M) \approx AN^{\alpha}M^{\beta}$ & 0.0044 \\
        $L(N, M) \approx AN^{\alpha}M^{\beta} + C$ & 0.0035 \\
        $L(N, M) \approx AN^{\alpha + \beta M} + C$ & \tabemphgood{0.0025} \\
        $L(N, M) \approx AN^{\alpha} + BM^{\beta} + C$ &  0.0043\\
        \bottomrule
    \end{tabular}
\end{table}

We see that the power law (row 1) and additive decomposition (row 4) are significantly worse at extrapolating loss values than more nuanced parametric forms. We note that the additive decomposition resembles the decomposition of loss as a function of model size $N$ and token budget $D$ used by \citet{hoffmann2022training}, but does not seem to reflect how $M$ affects loss for DiLoCo. We leave it as an open problem to determine what parametric forms better predict loss, and can be explained by theoretical understanding of communication-efficient training.

\section{Related Work}

\paragraph{Distributed training of LLMs.} Due to their increasingly large sizes, the advancement of language models has necessitated advancements in distributed training methods. One vein of work uses data-parallel training, and attempts to shard its constituent computations across accelerators in efficient ways. This includes advancements in things like distributed data parallelism~\citep{shazeer2018mesh,li2020pytorch}, ZeRO parallelism~\citep{rajbhandari2020zero, ren2021zero}, fully-sharded data parallelism~\citep{FairScale2021, zhao2023pytorch}, and pipeline parallelism~\citep{petrowski1993performance, huang2019gpipe, narayanan2019pipedream}. Conceptual models of the impact of batch size on training time~\citep{mccandlish2018empirical} aid in making the most effective trade-offs between training time and compute cost when using data-parallel training methods.

As scale continues to increase, the need to all-reduce gradients between data-parallel replicas becomes a bottleneck in training, causing accelerators to `wait' on this allreduce for an unacceptably long time. This observation has motivated extensive work in pipeline parallelism and scheduling, communicating activations rather than gradients. An alternative line of work keeps the basic data processing pattern of data parallelism while directly minimizing communication requirements.

Three broad families of algorithms exist in that space: 1) sparse updates (including CocktailSGD \citep{wang2023cocktailsgd}, PowerSGD \citep{vogels2020powersgd}, and DeMo \citep{peng2024decoupled}), 2) fast asynchronous updates (including Hogwild \citep{niu2011hogwildlockfreeapproachparallelizing}, WASH \citep{fournier2024washtrainensemblecommunicationefficient}, and Sparta \citep{exo2025sparta}), and 3) infrequent updates (including LocalSGD \citep{stich2018local}, FedOpt \citep{reddi2021adaptive}, and DiLoCo \citep{douillard2023diloco}).

In this work, we focus on the third category, which \citet{douillard2023diloco} proved recently it can reduce communication costs in LLM training significantly more by training multiple models independently with infrequent synchronization. Their method, DiLoCo, massively reduces communication overhead when training LLMs with a moderate numbers of model replicas. It has also shown great promise in training LLMs up to 10 billion parameters~\citep{jaghouar2024opendiloco,jaghouar2024intellect1technicalreport}. This work has also been extended to asynchronous overlapped updates \citep{liu2024asynchronous,douillard2025streaming}, and low-communication expert sharding \citep{douillard2024dipaco}.

\paragraph{Federated learning.} There is an enormous body of work on communication-efficient training methods for machine learning. In that vein, \dlc is closely related to algorithms used in federated learning to perform communication-efficient training over decentralized data, often (but not exclusively) on edge devices~\citep{kairouz2021advances}. The prototypical algorithm used in federated learning, FedAvg~\citep{mcmahan2017communication}, reduces communication costs by training models in parallel, with periodic model averaging. This algorithm has been invented and reinvented throughout machine learning, and is also known as Local SGD~\citep{stich2018local}, parallel SGD~\citep{zinkevich2010parallelized}, and parallel online backpropagation~\citep{mangasarian1993backpropagation}. The use of inner and outer optimization steps (as in \dlc, see \Cref{alg:diloco}) was first used by \citet{hsu2019measuring} and \citet{reddi2021adaptive} for federated learning, focusing on SGD as the inner optimizer and SGDM or Adam~\citep{kingma2014adam} as the outer optimizer in order to leverage more sophisticated optimizers in resource-constrained settings. \dlc is also closely related to many other federated optimization methods, though the huge amount of work in this area makes it impossible to summarize succinctly here. We instead refer the interested reader to the survey of~\citet{wang2021field}, though the field has of course progressed since then. While federated learning is often applied to more moderately sized models, \citet{charles2024towards} and \citet{sani2024photon} show that federated learning can be used to good effect for LLM training.

\paragraph{Scaling laws.} Scaling laws work often aims to estimate how empirical generalization error scales with various facets, including model size and training set size. Empirical scaling analyses with power law behavior date have existed for decades (see \citep{banko2001scaling}). \citet{hestness2017deep} developed power laws for model and dataset size across various tasks and model architectures (including encoder-decoder LSTM models). More recently, scaling laws for transformer-based LLMs were proposed by \citet{kaplan2020scaling} and \citet{hoffmann2022training}, who exhibited power law relationships between LLM performance and model size. Sine then, there has been a large number of works developing scaling laws for other facets of LLMs, including (among many others) inference costs~\citep{sardana2023beyond}, data-constrained training~\citep{muennighoff2023scaling}, and overtraining~\citep{gadre2024language}.

Scaling laws for \dlc were previously studied by \citet{he2024exploring}, who show that DiLoCo with 8 replicas exhibits analogous scaling behavior to \dpfull. \citet{he2024exploring} use a fixed number of replicas, batch size, and outer learning rate, and an unspecified total token budget.\footnote{At the time of writing, \citet{he2024exploring} only say that ``Each model was trained
to achieve adequate convergence''.} Our work expands on many aspects of their work and explores others that were not considered, including but not limited to: $10\times$ larger models, varying the number of replicas (including single-replica DiLoCo), varying token budgets and overtraining, parametric function fitting, scaling laws for hyperparameters, and optimal batch size analysis.

\section{Conclusions and Future Work}

Our results above all show that like \dpfull, \dlc scales predictably with model size in ways that make it simpler to tune hyperparameters and train models at extremely large scales. Moreover, \dlc can offer significant benefits over \dpfull, including superior evaluation loss when using a single model replica, and increased optimal batch size for any number of model replicas. Moreover, these benefits are robust to model scale, overtraining amount, and synchronization frequency.

There are at least three promising veins of future work in this space. First, the scaling law analyses can be augmented by various facets already used for \dpfull scaling laws. Examples include downstream task analysis~\citep{gadre2024language}, data-constrained scaling laws~\citep{muennighoff2023scaling}, and inference costs~\citep{sardana2023beyond}. Second, the scaling laws can be adapted to encompass proposed improvements to DiLoCo and related methods, including asynchronous updates~\citep{liu2024asynchronous}, streaming DiLoCo~\citep{douillard2025streaming}, and modular architectures co-designed with training methods~\citep{douillard2024dipaco,huh2024training}. Last, there is a clear need for systems and software that can be used to deploy methods like DiLoCo at scale and attain its communication-efficiency benefits in practical extremely large settings~\citep{jaghouar2024opendiloco}.

\section*{Acknowledgments}

We would like to thank Lechao Xiao for invaluable advice on scaling law methodology and using the NanoDO codebase; Adhiguna Kuncoro, Andrei Rusu, Jeffrey Pennington, Jiajun Shen, Mark Kurzeja, Nicole Mitchell, Qixuan Feng, Rachita Chhaparia, Ran Tian, Satyen Kale, Sean Augenstein, Vincent Roulet, Yani Donchev, and Zohar Yahav for helpful comments, advice, and conversations; Brendan McMahan, Daniel Ramage, Marc'Aurelio Ranzato, and Prateek Jain for feedback, guidance, and leadership support.

\bibliographystyle{plainnat}
\bibliography{ref}

\appendix

\section{Wall-Clock Time Model}\label{sec:wallclock_model}

In this section, we present an idealized model for the wall-clock times of \dpfull and \dlc. We measure the total elapsed time, which means that parallelization (e.g. via increasing the batch size) reduces wall-clock time.

\subsection{Computation Time}

Here, we mean the time expended by floating point operations in model training, ignoring communication time across nodes (which we treat separately in the section below). We use the idealized model where the total FLOPs $C = 6ND$. Given some number of chips $R$, each of which can perform $Q$ floating point operations per second, the total computation time is bounded below by $C / RQ$. The number of chips $R$ is purely a function of model size $N$ and global batch size $B$. The number of chips does not depend on the algorithm (\dpfull or \dlc) or number of model replicas when using \dlc.

\subsection{Communication Time}

The network connectivity is characterized by a bandwidth $W$ and latency $\epsilon$. When performing an all-reduce of $N$ parameters over $R$ compute nodes, the lower bound on the amount of traffic sent and received by at least one of the compute nodes participating in the all-reduce is $2N(1-R^{-1})$~\citep{Patarasuk2009Bandwidth}. Such algorithms are called \emph{bandwidth-optimal}. Since communication across the nodes is done synchronously but in parallel, in a network with bandwidth $W$ and latency $\eps$ between each pair of nodes, the time to complete the all-reduce is at least
\[
\dfrac{2 N}{W}\left(1 - \frac{1}{R}\right) +\eps.
\]

DiLoCo~\citep{douillard2023diloco} was designed for settings where models are too large to fit in a single datacenter, so they must be trained across compute islands connected by low bandwidth. To model this, we will assume that we are training over $R$ compute nodes (typically, GPUs or TPUs). Some of these are connected by networks within a datacenter, and others are connected across datacenters. We let $W_0, \eps_0$ denote the bandwidth and latency of the within-datacenter network, and $W_1, \eps_1$ analogously defined for the cross-datacenter network. Typically, $W_0 \geq W_1, \eps_0 \leq \eps_1$.

\paragraph{Data-Parallel:} At every training step $T$, we have to perform an all-reduce over all $R$ compute nodes. Since some nodes are connected across datacenters, the total communication time is at least
\[
\left(\dfrac{2N}{W_1}\left(1 - \frac{1}{R}\right) + \eps_1\right)T
\]

\paragraph{DiLoCo, $M = 1$:} At every inner step $T$, we perform an all-reduce over all $R$ devices as in Data-Parallel training. We also do an all-reduce every $H$ steps for the outer optimization. Some of these nodes are connected across datacenters, so the communication time per all-reduce is at least $2 N (1 - R^{-1})W_1^{-1} + \eps_1$. The total communication time is therefore at least
\[
\left(\dfrac{2N}{W_1}\left(1 - \frac{1}{R}\right)+ \eps_1\right)T\left(1 + \frac{1}{H}\right).
\]

\paragraph{DiLoCo, $M \geq 2$:} We assume that each of the $M$ model replicas is trained on compute nodes connected within the same datacenter. In each inner step $T$, each model replica is trained by $R/M$ devices which need to do an all-reduce. However, no communication occurs between datacenters, so the communication time of each inner step is bounded by $2 N (1 - MR^{-1})W_0^{-1} +\eps_0$.

Each outer optimization step involves all-reducing over all $R$ devices, connected across datacenters. This incurs a communication time of at least $2 N (1-R^{-1})W_1^{-1} + \eps_1$. Since it occurs only every $H$ steps, the total communication time is bounded below by:
\[
\left(\dfrac{2N}{W_0}\left(1 - \frac{M}{R}\right) + \eps_0\right)T + \left(\dfrac{2N}{W_1}\left(1 - \frac{1}{R}\right)+ \eps_1\right)\frac{T}{H}.
\]

Note that this suggests that as long as $H \geq W_0 / W_1$, the outer communication steps incur at most half of the total communication cost.

\paragraph{Streaming DiLoCo.} We note that the computed cost above applies to the Streaming DiLoCo \citep{douillard2025streaming} as well. While the inner step remains the same, the outer step is smoothed such that each fragment $p\,\in \{1, \dots, P\}$ is every $H$ steps. However, fragment communication is offset such that some fragment is communicated every $H / P$ steps, resulting in the communication amortizing to the same per-step cost. This is expected as Streaming DiLoCo reduces peak communication over any step, but does not reduce total communication across training.

\paragraph{Overlapping communications.} Another contribution of \citet{douillard2025streaming} is the ability to overlap communications required for the outer optimizer with computation by using a stale version of the fragment in the outer optimizer, and merging the result of this outer optimization with the locally optimized fragment. This would allow, for example, the communication term to be omitted from the calculation for wall-clock-time, if computation time dominates communication time. This setting is different from an algorithmic perspective, so its impact on scaling would need to be examined independently.

\subsection{Total Wall-Clock Time}

The total wall-clock time is a sum of the computation and communication times above. To measure the communication time, we must know the number of chips $R$ used for each experiment, the number of FLOPs per chip per second $Q$, the bandwidth and the latency of the within-datacenter and cross-datacenter networks. For computation costs, we use a slightly idealized number of chips $R$ based on our experiments, but ensuring that doubling the global batch size would double the number of chips. We base the constant $Q$ on publicly available information about the FLOPs capabilities of the TPU v5e and v6e chips\footnote{See https://cloud.google.com/tpu/docs/v6e.}, which have peak compute per chip (in bfloat16) of 197 teraflops and 918 teraflops, respectively. Assuming a maximum FLOPs usage of 50\%, these chips have an actual compute of approximately 100 and 408 teraflops, respectively. When computing idealized compute time, we set $Q = 300$ teraflops, somewhere in-between the two.

For bandwidth and latency, we consider three archetypes of networks:
\begin{enumerate}
    \item \textbf{High-bandwidth network} with bandwidth $\whigh = 400$ gigabits per second and a latency of $\epshigh = 10^{-4}$ seconds.
    \item \textbf{Medium-bandwidth network} with bandwidth $\wmed = 100$ gigabits per second and a latency of $\epsmed = 10^{-3}$ seconds.
    \item \textbf{Low-bandwidth network} with bandwidth $\wlow = 10$ gigabits per second and a latency of $\epslow = 10^{-2}$ seconds.
\end{enumerate}

We stress that these are not based on any actual systems, and are simply designed as instructive archetypes of networks. For the idealized communication time, we always use the high-bandwidth network for the within-datacenter network, and one of the three for the cross-datacenter network.

\clearpage

\section{Additional Experimental Results}

\begin{figure}[ht]
    \centering
    \includegraphics[width=\linewidth]{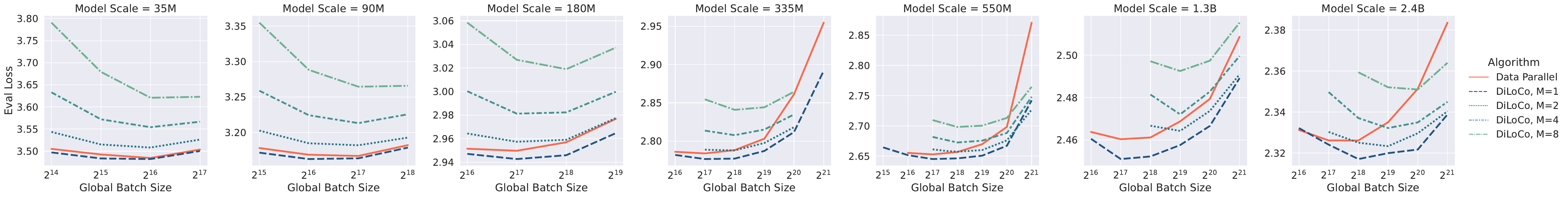}
    \caption{Evaluation loss of Data-Parallel and DiLoCo as a function of global batch size (in tokens).}
    \label{fig:batch_size_loss_appendix}
\end{figure}

\begin{figure}[ht]
    \centering
    \includegraphics[width=\linewidth]{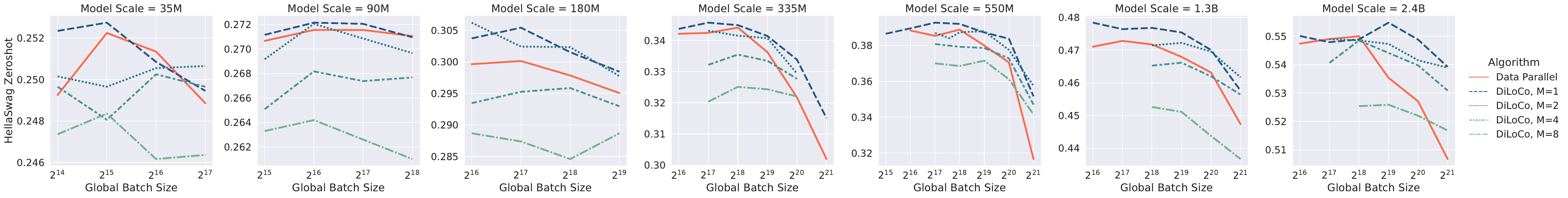}
    \caption{Zero-shot evaluation accuracy on HellaSwag of Data-Parallel and DiLoCo as a function of global batch size (in tokens).}
    \label{fig:batch_size_hellaswag_appendix}
\end{figure}

\begin{figure}[ht]
    \centering
    \includegraphics[width=\linewidth]{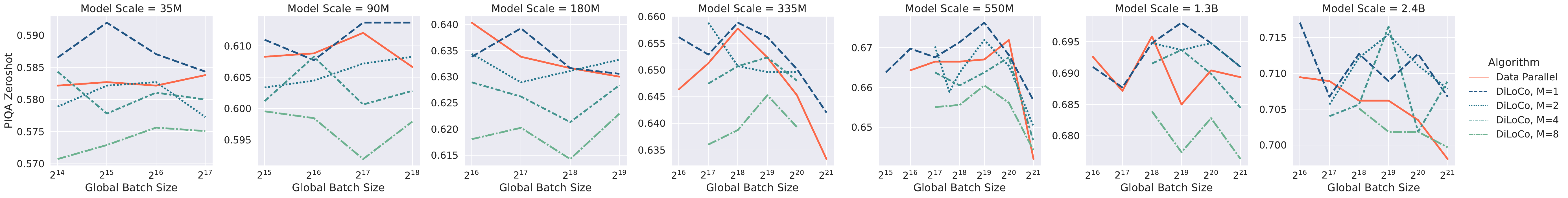}
    \caption{Zero-shot evaluation accuracy on Piqa of Data-Parallel and DiLoCo as a function of global batch size (in tokens).}
    \label{fig:batch_size_piqa_appendix}
\end{figure}

\begin{figure}[ht!]
    \centering
    \includegraphics[width=\linewidth]{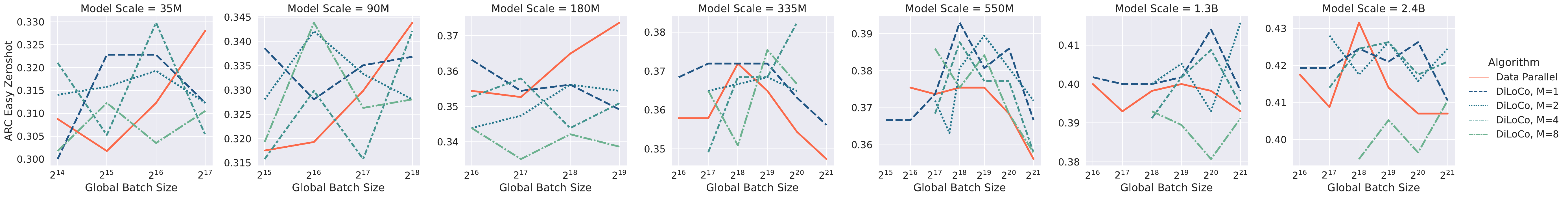}
    \caption{Zero-shot evaluation accuracy on Arc-Easy of Data-Parallel and DiLoCo as a function of global batch size (in tokens).}
    \label{fig:batch_size_arceasy_appendix}
\end{figure}

In this section, we give additional experimental results that expand on those in \Cref{sec:exp_results}. In Figures \ref{fig:batch_size_loss_appendix}, \ref{fig:batch_size_hellaswag_appendix}, \ref{fig:batch_size_piqa_appendix}, and \ref{fig:batch_size_arceasy_appendix}, we present evaluation loss and evaluation accuracy on various downstream zero-shot tasks, as a function of algorithm, model size, and global batch size. The results consistently show that \dpfull's evaluation performance degrades quickly as batch size increases. By contrast \dlc's performance degrades more slowly, or even improves, as batch size increases. We note that Arc-Easy was quite noisy as an evaluation task.

\begin{figure}[htb]
    \centering
    \begin{subfigure}[b]{0.48\linewidth}
    \includegraphics[width=\linewidth]{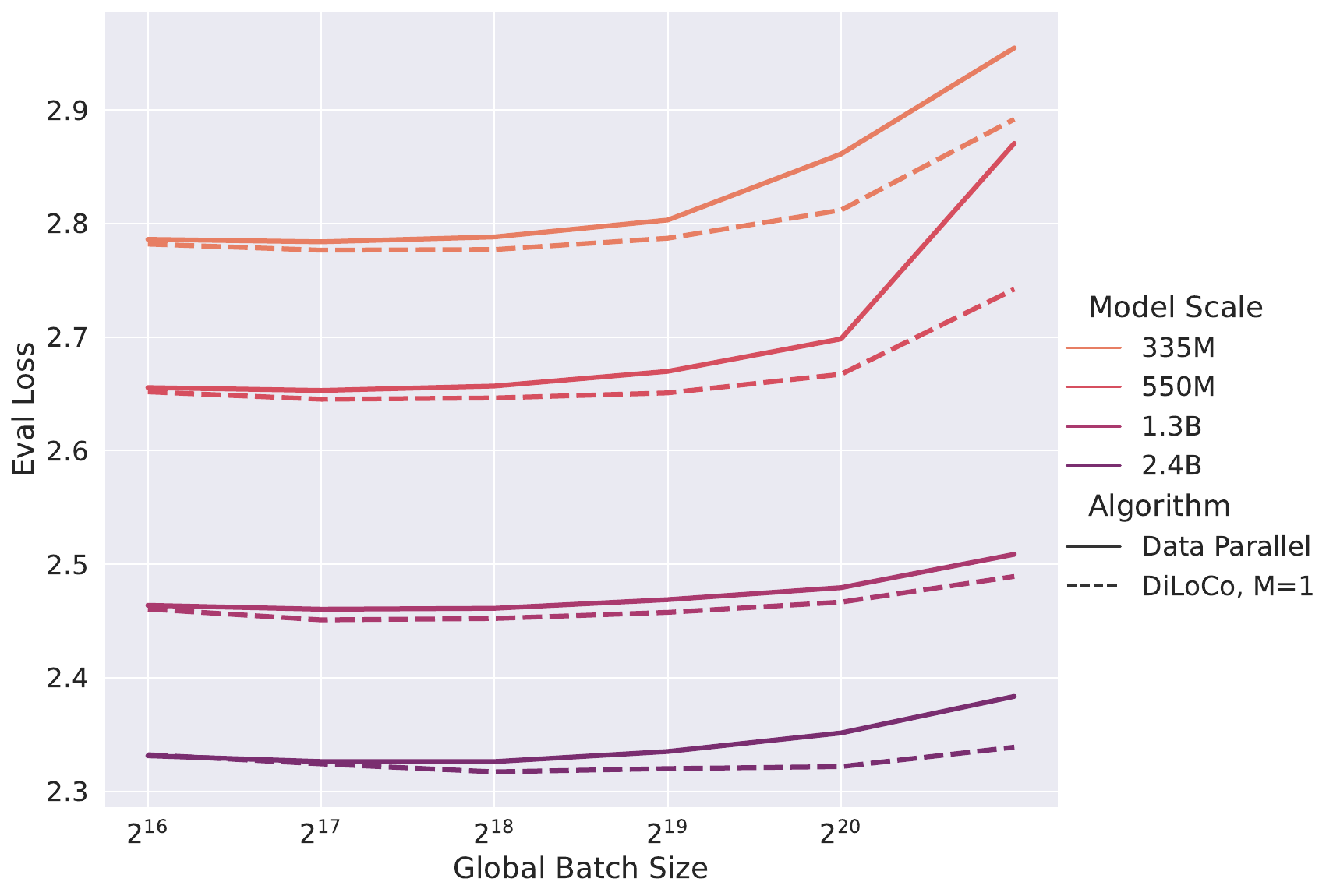}
    \caption{Evaluation loss.}
    \end{subfigure}
    \begin{subfigure}[b]{0.48\linewidth}
    \includegraphics[width=\linewidth]{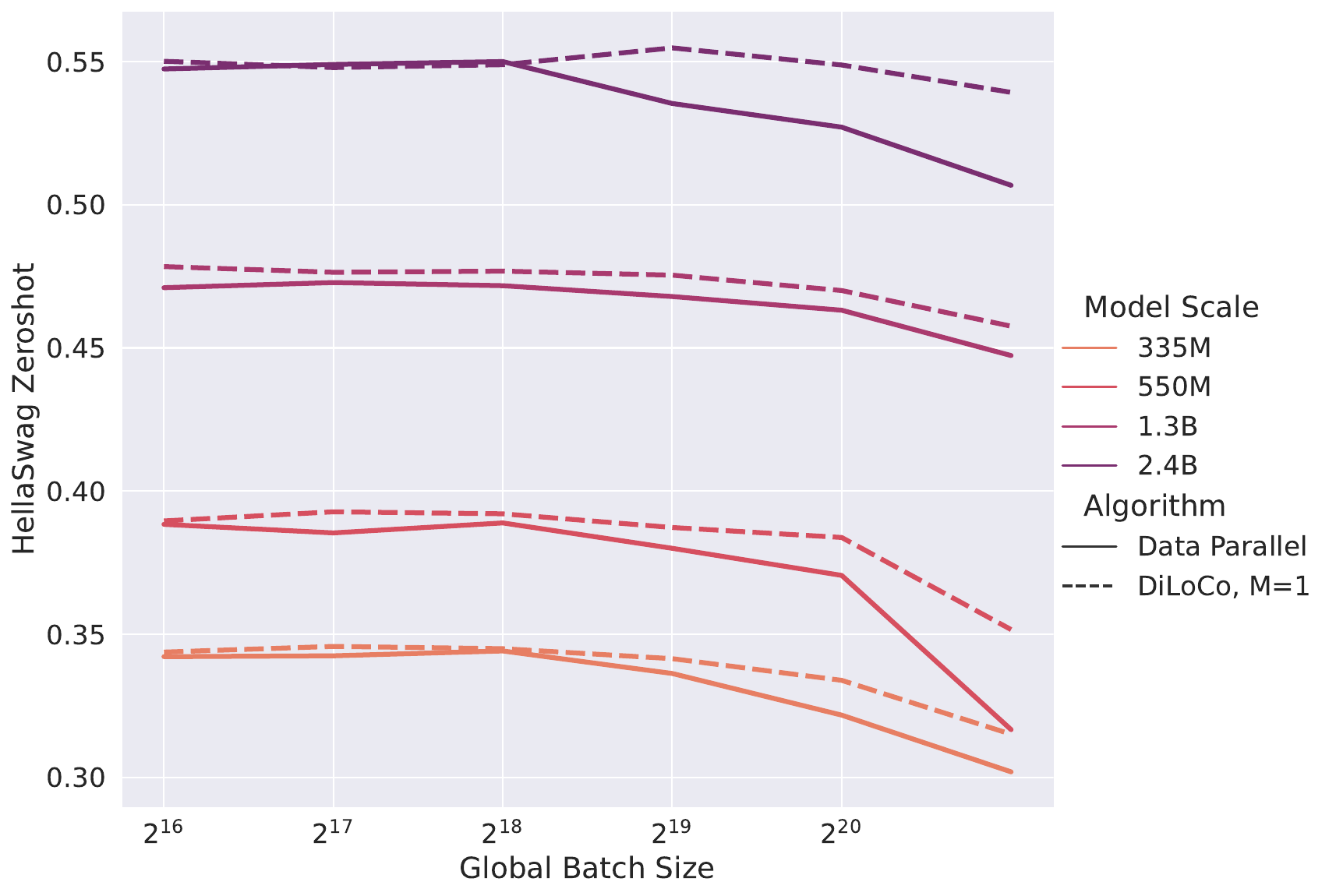}
    \caption{Zero-shot accuracy on HellaSwag.}
    \end{subfigure}
    \caption{Evaluation loss and zero-shot accuracy of \dpfull and DiLoCo with $M = 1$ for varying model and global batch sizes (measured in tokens). In all settings, DiLoCo with $M = 1$ does better than \dpfull, and the gap between them increases with batch size.}
    \label{fig:compare_dp_m1_main_alt}
\end{figure}

\begin{figure}[htb]
    \centering
    \begin{subfigure}[b]{0.48\linewidth}
    \includegraphics[width=\linewidth]{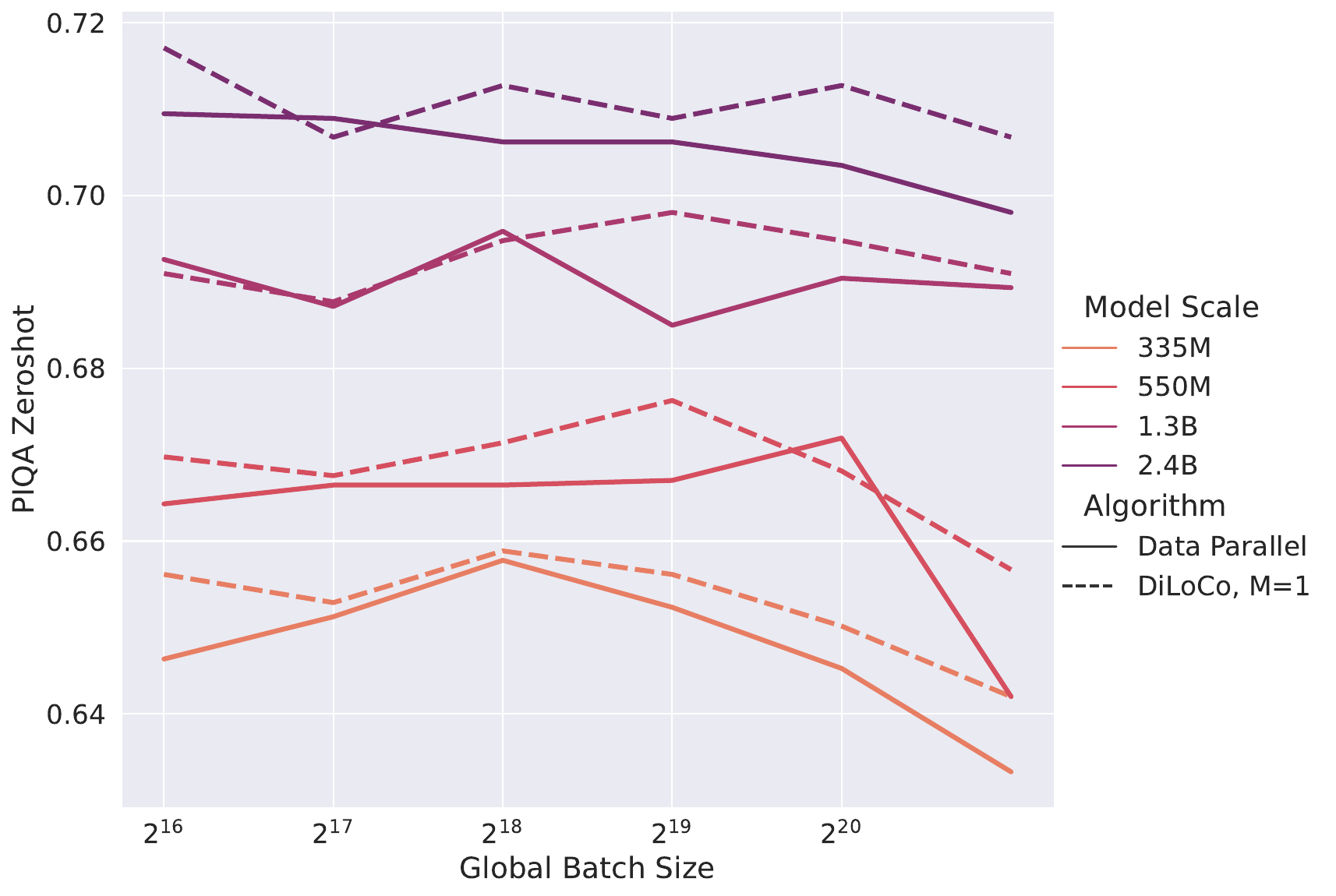}
    \caption{Zero-shot accuracy on Arc-Easy.}
    \end{subfigure}
    \begin{subfigure}[b]{0.48\linewidth}
    \includegraphics[width=\linewidth]{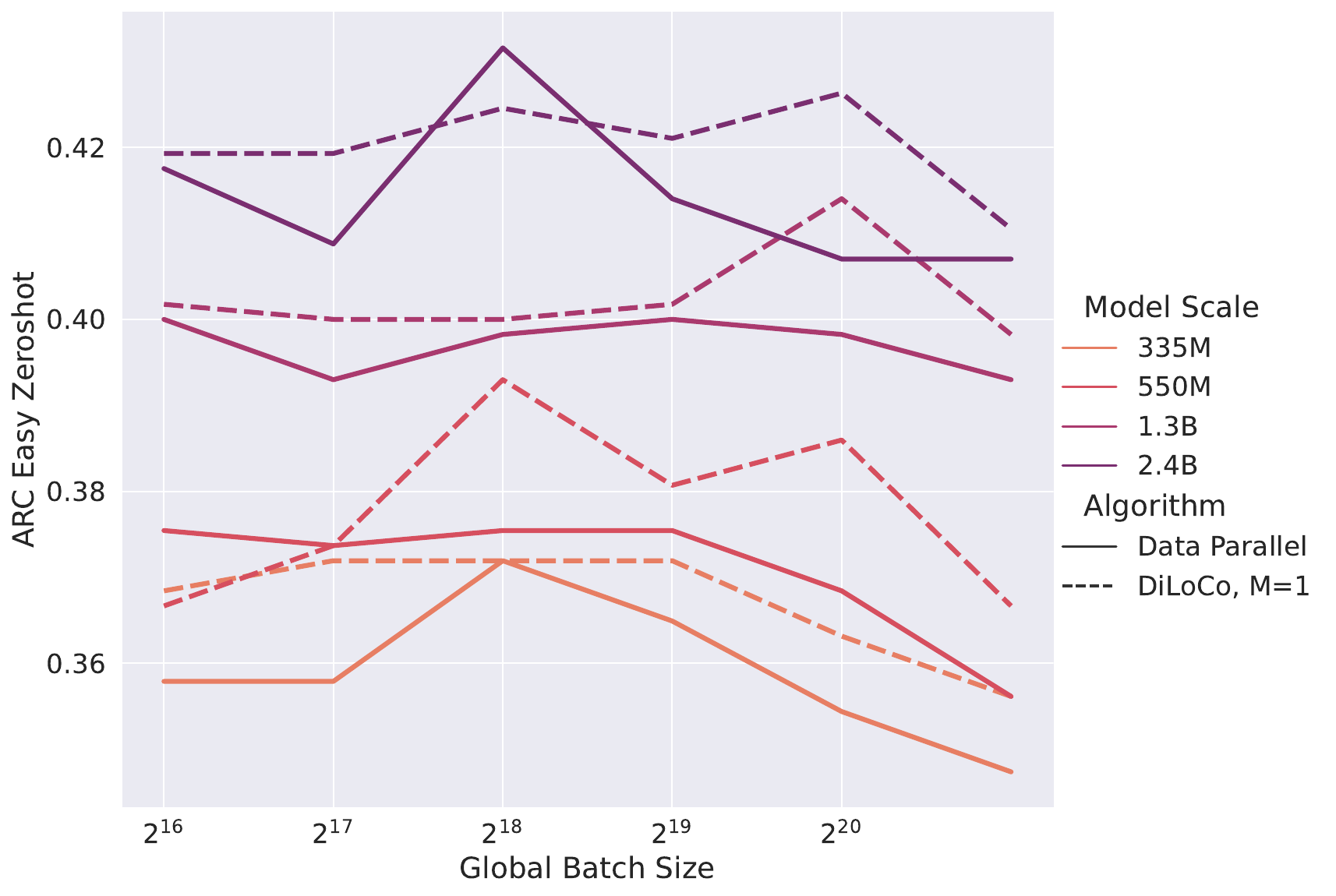}
    \caption{Zero-shot accuracy on Arc-Easy.}
    \end{subfigure}
    \caption{Zero-shot evaluation accuracy of \dpfull and DiLoCo with $M = 1$ for varying model and global batch sizes (measured in tokens), on Piqa and Arc-Easy. In nearly all settings, DiLoCo with $M = 1$ does better than \dpfull, and the often the gap increases with batch size.}
    \label{fig:compare_dp_m1_appendix}
\end{figure}

In Figures \ref{fig:compare_dp_m1_main_alt} and \ref{fig:compare_dp_m1_appendix}, we compare \dpfull and \dlc with $M = 1$ in terms of their evaluation loss and zero-shot evaluation accuracy on HellaSwag, Piqa and Arc-Easy. As above, we note that \dlc with $M = 1$ has an improved tolerance to larger batch sizes.

\end{document}